\documentclass{article}
\usepackage[numbers]{natbib}
\usepackage{graphicx} 
\usepackage{tabularx}
\usepackage{pifont}
\usepackage[table]{xcolor}

\newcommand{\spade}{\textcolor{black}{\ding{171}}}   
\newcommand{\heart}{\textcolor{red}{\ding{170}}}     
\renewcommand{\diamond}{\textcolor{red}{\ding{169}}}   
\newcommand{\club}{\textcolor{black}{\ding{168}}}    

\definecolor{wincolor}{RGB}{198,239,206}
\definecolor{drawcolor}{RGB}{255,235,156}
\definecolor{losecolor}{RGB}{255,199,206}

\newcommand{\wdlwin}[1]{\cellcolor{wincolor}#1}
\newcommand{\wdldraw}[1]{\cellcolor{drawcolor}#1}
\newcommand{\wdllose}[1]{\cellcolor{losecolor}#1}

\usepackage[preprint]{neurips_2025}

\usepackage[utf8]{inputenc} 
\usepackage[T1]{fontenc}    
\usepackage{hyperref}       
\usepackage{url}            
\usepackage{booktabs}       
\usepackage{amsfonts}       
\usepackage{nicefrac}       
\usepackage{microtype}      
\usepackage{xcolor}         
\usepackage[breakable]{tcolorbox}
\usepackage{algorithm}
\usepackage{algpseudocode}
\usepackage{amsmath}
\usepackage{multirow}
\usepackage{newunicodechar}
\usepackage{makecell}
\newunicodechar{◁}{\textlangle}
\newunicodechar{▷}{\textrangle}
\usepackage{listings}
\title{Divide-Fuse-Conquer: Eliciting ``Aha Moments'' in Multi-Scenario Games}

\author{
\begin{tabular}{c}
Xiaoqing Zhang$^{1,2}$\thanks{This work was done during the internship at Moonshot AI.} \quad \quad 
Huabin\  Zheng$^{2}$ \quad \quad 
Ang\  Lv$^{1}$ \quad \quad 
Yuhan\  Liu$^{1}$ \quad \quad 
Zirui Song$^{3}$ \quad \quad \\
Xiuying Chen$^{3}$\footnotemark[2] \quad \quad \ \ Rui \ Yan$^{1}$\thanks{\ \ Corresponding authors.} \quad \quad Flood Sung$^{2}$ 
\end{tabular}
\\ \vspace{.5mm}
    \small
    \begin{tabular}{c}
    $^1$Gaoling School of Artificial Intelligence, Renmin University of China \quad $^2$Moonshot AI\\ 
    $^3$Mohamed bin Zayed University of Artificial Intelligence\\
    \end{tabular}
    \\ \vspace{.5mm}
    \small
    \begin{tabular}{c}
    \texttt{\{xiaoqingz, anglv, yuhan.liu, ruiyan\}@ruc.edu.cn} \quad \texttt{zirui.song@mbzuai.ac.ae}\\ \texttt{\{watson, floodsung\}@moonshot.cn} \quad \texttt{xy-chen@pku.edu.cn}\\
    \end{tabular}
    \vspace{2mm} \\
}

\begin{document}

\maketitle

\begin{abstract}
Large language models (LLMs) have been observed to suddenly exhibit advanced reasoning abilities during reinforcement learning (RL), resembling an ``aha moment'' triggered by simple outcome-based rewards.
While RL has proven effective in eliciting such breakthroughs in tasks involving mathematics, coding, and vision, it faces significant challenges in multi-scenario games.
The diversity of game rules, interaction modes, and environmental complexities often leads to policies that perform well in one scenario but fail to generalize to others.
Simply combining multiple scenarios during training introduces additional challenges, such as training instability and poor performance.
To overcome these challenges, we propose Divide-Fuse-Conquer, a framework designed to enhance generalization in multi-scenario RL. 
This approach starts by heuristically grouping games based on characteristics such as rules and difficulties. 
Specialized models are then trained for each group to excel at games in the group is what we refer to as the divide step.
Next, we fuse model parameters from different groups as a new model, and continue training it for multiple groups, until the scenarios in all groups are conquered.
Experiments across 18 TextArena games show that Qwen2.5-32B-Align trained with the Divide-Fuse-Conquer strategy reaches a performance level comparable to Claude3.5, achieving 7 wins and 4 draws.
We hope our approach can inspire future research on using reinforcement learning to improve the generalization of LLMs.
\end{abstract}

\begin{figure*}[htb]
  \centering
  \includegraphics[width=0.8\textwidth]{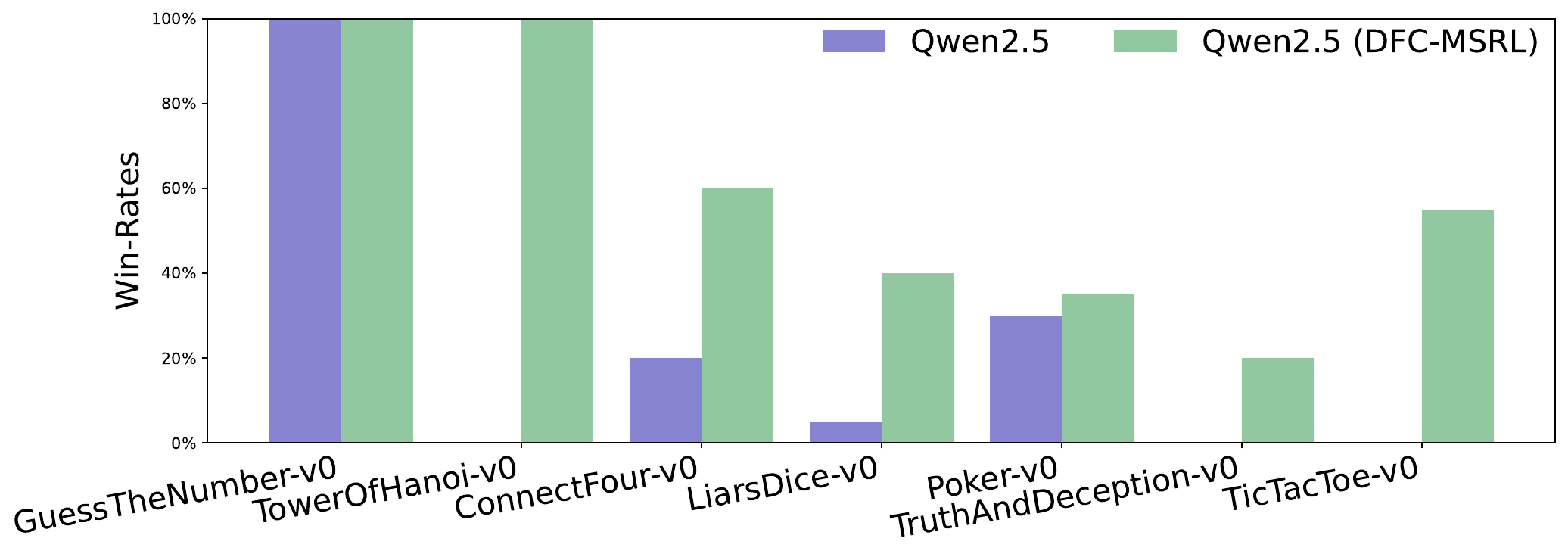}
  \caption{The win rates of Qwen2.5 and Qwen2.5 (DFC-MSRL). Qwen2.5 refers to the base model Qwen2.5-32B-Align, while Qwen2.5 (DFC-MSRL) denotes the fine-tuned model using the Divide-Fuse-Conquer strategy. For single-player games such as GuessTheNumber-v0 and TowerOfHanoi-v0, the win rate represents the success of interacting with the environment. For all other two-player games, the win rate indicates the frequency with which the model outperforms Claude3.5.}
  \label{fig:data}
\end{figure*}


\section{Introduction}
In recent years, LLMs have demonstrated impressive reasoning abilities, exhibiting interpretable interaction patterns and growing proficiency in strategic decision-making across domains such as math, coding, and vision tasks~\cite{shao2024deepseekmath,xie2025logic,shen2025satori,guo2024deepseek,huang2025vision}.
Building on these successes, games have emerged as a natural next domain for evaluating and advancing LLM capabilities, offering rich, interactive environments that require agents to plan, adapt, and reason under uncertainty.
Unlike single-scenario settings such as math or coding, games require agents to operate across diverse mechanics and shifting objectives, demanding both flexible decision-making and strong generalization across scenarios~\cite{kanagawa2019rogue,xu2021generalization,qi2024civrealm}.
This makes games an ideal testbed for exploring the integration of RL with LLMs, presenting both exciting opportunities and significant challenges~\cite{sweetser2024large,ma2024coevolving,singh2025agentic,alabi2025reinforcement,baek2025pcgrllm,zhang2025r1}.
DeepSeek-R1~\cite{guo2025deepseek} introduces a novel paradigm that eliminates the need for SFT data, relying solely on reinforcement learning to develop long-chain reasoning and reflective capabilities. 
Building on this foundation, we explore the potential of using R1 and find that as the number and heterogeneity of game scenarios increase, we observe severe issues such as training collapse and poor performance, which limit R1's applicability to multi-scenario games. 
These issues stem from its inability to reliably generalize across diverse scenarios and the lack of supervised guidance to stabilize learning.

To address these challenges, we propose a Divide-Fuse-Conquer strategy. In this framework, games are first heuristically grouped based on characteristics such as game rules and difficulties—this is the divide step. 
Specialized models are trained to perform well within each group.
Next, we initiate the fuse phase by merging model parameters from different groups into a new model, which is then further trained to conquer the combined set of scenarios. 
We repeat this fuse and conquer process progressively until the model has been trained on all groups.
The core of our approach lies in the conquer phase, where we focus on optimizing performance by refining the training process to effectively consolidate and expand the knowledge learned from different game groups.
To achieve this, we introduce a suite of game-specific techniques aimed at enhancing the stability, efficiency, and performance of training among groups.
For stability, we incorporate format reward shaping to ensure valid action for environments and apply half-negative sampling to prevent training collapse by reducing the influence of overwhelming negative gradients.
To improve efficiency, we adopt mixed priority sampling, which assigns higher sampling probabilities to trajectories from games with low or intermediate win rates. 
This encourages learning in scenarios where the model has not yet converged or remains uncertain.
To enhance performance, we employ an $\epsilon$-greedy disturbance~\cite{cesa1998finite} along with a randomized seed to diversify the distribution of actions and initial states, encouraging broader exploration. 
Additionally, we introduce a step reward called a hasty-action penalty to discourage impulsive decisions and guide the model toward deeper, more strategic reasoning.

To evaluate our method in gaming scenarios, we conducted experiments across 18 games of varying types and difficulty levels within the TextArena~\cite{guertler2025textarena}, using the Qwen2.5-32B-Align model.
Despite the strong performance of the Claude3.5 baseline, our method successfully learned to play these games and demonstrated competitive results, achieving head-to-head records up to W7:D4:L7 (where W, D, and L represent Wins, Draws, and Losses, respectively).

In summary, our contributions are as follows:

$\bullet$ We propose Divide-Fuse-Conquer, a progressive training framework designed to enhance training stability in multi-scenario reinforcement learning.
This framework addresses the challenge of training collapse caused by heterogeneous task distributions across games.

$\bullet$ We integrate a suite of techniques to improve the stability, efficiency, and performance of multi-scenario training, including format reward shaping, half-negative sampling, $\epsilon$-greedy disturbance, randomized seed initialization, mixed priority sampling, and the step reward shaping by hasty-action penalty.
These techniques mitigate issues such as poor performance in simple games, slow convergence at the beginning, and shallow reasoning in multi-step scenarios.

$\bullet$ We conduct extensive experiments across 18 diverse games in TextArena and provide in-depth analysis of model behaviors during training. 
Our findings offer insights into LLM-based RL, guiding future efforts to scale it for more robust and generalizable multi-scenario games.

\section{Related works}

\textbf{Reinforcement learning for LLMs.}
Recent work has increasingly explored RL as a means to enhance the reasoning and alignment capabilities of LLMs across diverse domains~\cite{guo2025deepseek,huang2025vision,zhou2025r1}.
~\citeauthor{rafailov2023direct} introduced Direct Preference Optimization, which analytically maximizes the likelihood of human preferences without relying on a reward model or policy sampling—offering a streamlined alternative to conventional RLHF.
Building on this, ~\citeauthor{Amini2024} proposed Offset-DPO, a method that assigns greater weight to stronger preferences, thereby improving alignment under limited preference data.
These alignment-focused innovations complement domain-specific RL approaches aimed at enhancing reasoning.
In mathematical reasoning, ~\citeauthor{shao2024deepseekmath} presented DeepSeekMath, which integrates continued pretraining with a novel optimization method called Group Relative Policy Optimization (GRPO). This approach achieved strong performance on the MATH benchmark, demonstrating RL's effectiveness in symbolic reasoning.
In code generation, ~\citeauthor{Dutta2024} introduced RL4Code, which uses GPT-3.5 to generate preference signals and trains smaller models via RLAIF. This method significantly improves code executability and outperforms larger fine-tuned baselines.
In multimodal settings, ~\citeauthor{zhai2024fine} proposed Reasoning with Actions, a framework in which vision-language models generate intermediate reasoning steps and refine their policies through interaction with the environment—achieving state-of-the-art results in complex visual reasoning tasks.
Collectively, these works underscore the growing potential of RL and preference-based learning techniques in advancing the alignment and reasoning capabilities of LLMs across a wide range of modalities and tasks.

\textbf{Reinforcement learning in strategy-based games.}
The integration of reinforcement learning and large language models has seen significant advancements in gaming applications~\cite{shaheen2025reinforcement,amato2010high,liu2021self}.
Early research primarily employed LLMs to generate reward functions and game configurations that support RL training.
For example,\citeauthor{kwon2023reward} used LLMs to design reward functions for the Deal or No Deal negotiation game, based on its objectives and rules. 
These functions guided agents toward behaviors aligned with the game's goals by capturing both success and failure signals.
Similarly,~\citeauthor{todd2023level} explored using LLMs to generate levels for Sokoban, enhancing agents’ robustness and adaptability across diverse scenarios.
As the synergy between RL and LLMs deepened, researchers began moving beyond treating LLMs as mere auxiliary tools, instead fostering mutual reinforcement between the two.
For instance,~\citeauthor{xu2023language} introduced a framework where LLMs generated strategic suggestions for RL agents based on a deep understanding of the rules in Werewolf. 
The agents tested these strategies, and the results were fed back into the LLMs for iterative refinement.
In another example,~\citeauthor{tanaka2025grammar} combined SFT with GRPO to iteratively improve agents' ability to generate accurate game descriptions.
In multi-agent or cooperative game environments, maintaining training stability presents a significant challenge.
To address this,~\citeauthor{shi2024sample} proposed a sample-efficient, model-based algorithm for learning robust strategies in distributionally robust Markov games. This method aims to sustain multi-agent performance under high environmental uncertainty.

\section{Methods}
\subsection{Base Model}
We use Qwen2.5-32B-Align, an in-house model that is generally aligned from Qwen2.5-32B-Base through a multi-stage alignment process.
In the first stage, the model underwent Supervised Fine-Tuning (SFT) on large-scale annotated data to strengthen its core capabilities and language understanding.
In the second stage, we applied Direct Preference Optimization (DPO)~\cite{rafailov2023direct}, using human preference data to enhance response quality and better align outputs with human expectations.
This two-stage process equips Qwen2.5-32B-Align with strong general task performance, precise adherence to formatting instructions, and the ability to generate safe and reliable outputs across diverse languages and application scenarios.

\subsection{Training Recipe}

At the beginning, R1 is effective in single-scenario games. When extended to multi-scenario settings, we observed its strong performance on similar games, even enabling some that had previously shown no improvement with individual training to achieve significant gains. However, when the games were excessively dissimilar, the training became unstable. These observations suggest that both single-scenario training and multi-scenario training involving too many or highly diverse games are suboptimal.
Nonetheless, some tasks, when trained individually, are completely unable to break away from zero performance. This indicates that while single-scenario training can be effective for certain tasks, it may not be sufficient for all, especially those that require more diverse training environments to improve.
As a result, we are motivated to select a more focused set of similar tasks for training to enhance the stability of the learning process.
To address these challenges, we adopt a Divide-Fuse-Conquer strategy.
Our method is formally described in Algorighm~\ref{alg:gptMtrl}.

\begin{algorithm}[tb]
\small
\caption{Divide-Fuse-Conquer}
\label{alg:gptMtrl}
\begin{algorithmic}[1]

\State Initialize the game set $G$ that consists of multiple single-player and two-player games
\State Divide $N$ games $[g_1,g_2,g_3,...,g_N]$ into $K$ groups $G = \{ \mathcal{G}_k \}_{k=1}^K$ according to game characteristics
\State Initialize the policy $\pi$
\For {each $k$ in range(1, $K+1$)}
    \State Train on group $\{ \mathcal{G}_k \}_{k=k}^k$ with conquer recipe~\ref{alg:steprl} based on $\pi$, obtaining the policy $\pi_{\{ \mathcal{G}_k \}_{k=k}^k}$
\EndFor
\State Set the best policy $\pi'=\pi_{\{ \mathcal{G}_1 \}_{k=1}^1}$
\For {each $k$ in range(2, $K+1$)}
    \State Fuse the group policy and the best policy to obtain the fused model: $\pi' = Fuse(\pi_{\{ \mathcal{G}_k \}_{k=k}^k} + \pi')$
    \State Train on group $\{ \mathcal{G}_k \}_{k=1}^k$ with conquer recipe based on $\pi'$, obtaining the policy $\pi'_{\{ \mathcal{G}_k \}_{k=1}^k}$
    \State Set the best policy $\pi' = \pi'_{\{ \mathcal{G}_k \}_{k=1}^k}$
\EndFor
\State Return the best policy $\pi'$
\end{algorithmic}
\end{algorithm}

\subsubsection{Divide: grouping games by scenario characteristics}
In multi-scenario game RL, variations in game mechanics, interaction modes, and environmental complexity present significant challenges to achieving stable and efficient training. 
The 18 games in TextArena exhibit wide diversity in both rules and difficulties. 
Some games start from a fixed initial state with identical observable information and conditions for all players (e.g., ConnectFour-v0, TicTacToe-v0, TowerOfHanoi-v0), while others begin with varied configurations, such as private hands or asymmetric observations (e.g., Poker-v0, LiarsDice-v0).
When starting from the same initial state, the model faces a limited observation and action space, which restricts exploration and increases the risk of getting trapped in local optima compared to using randomized starting states. For easier games, the model tends to reinforce high-reward behaviors, effectively learning within its comfort zone. Since optimizing these high-reward behaviors typically yields faster returns, training with easier games and harder games together often leads the model to prioritize the former, thereby slowing learning on the latter.
To ensure a consistent learning pace within each group, we divide the games into four distinct categories based on the rules and difficulties of the game.
Specifically, we first classify the games into two categories: those with random initial states and those with fixed initial states. Within each category, games are further divided based on whether the base model achieves a non-zero win rate—that is, whether it can win at all. This distinction serves as a proxy for game difficulty, yielding a total of four groups.

\subsubsection{Fuse: integrating specialized policies}
In the multi-scenario setting, we assume that each group $\mathcal{G}_k$ consists of games that share similar features. 
For each group, we first train an individual model-a policy denoted as $\pi_k$, with parameters $\theta_{\pi_k}$-which is capable of capturing the intragroup dynamics effectively. 
However, due to the diversity between groups, a policy trained solely within one group may not generalize well to others.
To address this challenge and enable knowledge transfer across groups, we introduce a policy fusion mechanism. Specifically, we maintain a fused policy, denoted $\pi^{(k-1)}$, which integrates knowledge from the first $k-1$ groups.
Its parameters $\theta^{(\pi^{(k-1)})}$ are obtained by averaging over previously learned policies. 
Once the individual policy $\pi_k$ is obtained for group $\mathcal{G}_k$, we combine it with the previously fused policy $\pi^{(k-1)}$ via parameter averaging.
This results in an updated fused policy $\pi^{(k)}$ with parameters $\theta^{(\pi^{(k)})} = \frac{1}{2} \left( \theta^{(\pi^{(k-1)})} + \theta_{\pi_k} \right)$. Here, $\pi^{(k)}$ denotes the new fused policy after incorporating the k-th individual model $\pi_k$. 

This fusion allows the newly acquired policy to inherit useful knowledge from previous groups, while also contributing its specialized understanding of group $\mathcal{G}_k$. 
The result is a gradually evolving policy that balances both stability and adaptability. 
By enhancing cross-group fusion, this mechanism increases the likelihood of generating high-quality trajectories across diverse groups, which facilitates faster generalization during continued training.

\subsubsection{Conquer: generalizing across groups}
After fusing models among groups, the fused model exhibits a parameter-shift compared to the models within each group. 
To generalize effectively to the fused scenario, we perform continuous training with RL on the merged group. 
To enhance the stability, efficiency, and overall performance of this process, we integrate a suite of techniques with RL. 

\textbf{Reward Shaping}
Reward modeling is a critical component of reinforcement learning, as it directly influences both learning efficiency and overall performance.
We design a rule-based reward system in Table~\ref{tab:reward} that assigns rewards to trajectory $\tau$, consisting of format rewards, environment rewards, and hasty-action penalties.
The format reward $R_{\text{format}}(\tau)$ is designed to encourage the model to follow instructions and produce executable actions.
Responses that fail to adhere to the specified format are penalized with a reward of -2.
The environment reward $R_{\text{env}}(\tau)$ is determined by the game outcome in TextArena, assigning -1 for a loss, 0 for a draw, and 1 for a win.
The hasty-action penalty, $R_{\text{step}}(\tau)$, is designed to improve the LLM’s decision-making by discouraging inefficient strategies and promoting faster victories. 
To better incentivize optimal behavior, we scale rewards based on the trajectory steps: for instance, solving the Tower of Hanoi after several ineffective moves yields a lower reward than solving it efficiently. 
Specifically, we use the number of steps $n_{\tau}$ in trajectory $\tau$ as a penalty factor in winning scenarios.

\begin{table}[htbp]
  \small
  \caption{Definitions of the format reward, environment reward, and step penalty reward. Here, $n_{\tau}$ denotes the number of steps in trajectory $\tau$.}
  \resizebox{0.9\linewidth}{!}{
    \begin{tabular}{ccc}
    \toprule
    \multicolumn{1}{c}{\textbf{Format Reward}} & \multicolumn{1}{c}{\textbf{Environment Reward}} & \multicolumn{1}{c}{\textbf{Step Penalty Reward}} \\
    \midrule
    $ R_{\text{format}}(\tau) = \begin{cases}-2, & \text{invalid} \\ 0, & \text{valid} \end{cases} $     & $ R_{\text{env}}(\tau) = \begin{cases}1, & \text{Win} \\ 0, & \text{Draw} \\ -1, & \text{Lose} \end{cases} $     & $R_{\text{step}}(\tau) = 
\begin{cases} 
R_{\text{format}}(\tau), & \text{if } R_{\text{format}}(\tau) < 0 \\
\frac{R_{\text{env}}(\tau)}{n_{\tau}}, & \text{if } R_{\text{env}}(\tau) > 0 \\
R_{\text{env}}(\tau), & \text{otherwise}
\end{cases}$ \\
    \bottomrule
    \end{tabular}}%
  \label{tab:reward}%
\end{table}%

\textbf{Half-negative}
We mitigate training collapse by selectively filtering negative trajectories during reinforcement learning. 
Specifically, we randomly discard half of the negative samples that may produce disproportionately large gradients. 
This prevents the optimization process from being dominated by negative samples and promotes more stable updates.

\textbf{Curriculum-guided Sampling} 
Given that each group $\mathcal{G}_k$ includes games of different difficulty levels, uniformly sampling data during multi-scenario RL can introduce inefficiencies, as varying learning speeds may disrupt balanced optimization across tasks.
Inspired by curriculum learning, we design a Mixed Prioritized Sampling (MPS) strategy that prioritizes games with the highest uncertainty and the lowest win rates for more frequent training.
This approach helps regulate the learning pace across tasks of varying difficulty.
Based on each game's historical win rate, MPS dynamically adjusts the number of seeds per game in each iteration $t$ by assigning different sampling probabilities.
To ensure that both low- and moderate-win-rate games in $\mathcal{G}_k$ are adequately trained, we combine the linear-capped and variance priority into a mixed sampling weight for each game $g_i$ in $\mathcal{G}_k$ at iteration $t$, denoted as $\text{W}_{g_i}^{(t)} = a \times \max(\epsilon_1, 1 - \text{WR}_{g_i}^{(t-1)}) + b \times \text{WR}_{g_i}^{(t-1)} \times (1 - \text{WR}_{g_i}^{(t-1)})$, where the $\text{WR}_{g_i}^{(t-1)}$ represents the win rate of game ${g_i}$ at iteration $t-1$.
$a$ and $b$ represent the weight of linear-capped priority and the variance priority, while $\epsilon_1$ represents the capped ratio, which defines the minimum sampling proportion for each game.
With the whole rollout budget $S$, we assign the rollout seeds of each game $g_i$ at iteration $t$ as $\text{S}_{g_i}^{(t)} = S \times \frac{\text{W}_{g_i}^{(t)}}{\sum_{j=1}^{|\mathcal{G}_k|} \text{M}_{g_j}^{(t)}}$.
With MPS strategy, the number of rollout seeds $S_{g_i}^{(t)}$ for each game $g_i$ is dynamically adjusted according to its win rate $\text{WR}_{g_i}^{(t-1)}$ during the last iteration, ensuring that the model focuses more on games with lower and moderate win rates, thus improving learning efficiency.

\textbf{$\epsilon$-greedy Disturbance}
We inject a small degree of randomness into the action selection process by occasionally replacing the model’s chosen action with a randomly sampled one. 
This controlled exploration helps the model escape local optima and discover more diverse and effective strategies.

\textbf{Randomized Seed Initialization}
We randomize environment seeds during training to expose the model to a wide variety of initial states and dynamics. 
This prevents overfitting to specific trajectories and enhances the generalization ability across different game configurations.

\textbf{Conquer Recipe} 
We conduct self-play rollouts to explore the application of multi-scenario reinforcement learning in games.
In single-player games, our model operates as Player 0, interacting directly with the environment. 
Conversely, in two-player games, the model alternates its role between Player 0 and Player 1 across different episodes. 
Each rollout involves playing the game to completion, with moves determined by the model acting in the designated player role. 
Given a rollout budget $S$, we accumulate a diverse set of trajectory data by sampling from the current best historical policy.
This data then serves as the foundation for learning via the GRPO~\cite{shao2024deepseekmath} algorithm. 
Furthermore, the aforementioned game-specific techniques are incorporated during self-play reinforcement learning to optimize the model’s behavior.
To ensure continuous improvement, we maintain a record of the policy exhibiting the highest performance. 
This best-performing policy then becomes the target for the subsequent iteration's update, determined by evaluating the win rate of the newly updated policy against the initial model.
The detailed training procedure is outlined in the algorithm~\ref{alg:steprl}.

\begin{algorithm}[tb]
\small
\caption{Conquer recipe}
\label{alg:steprl}
\begin{algorithmic}[1]

\State Initialize the group of training games $\mathcal{G}_k$, the training iterations $T$, the number of rollout seed for each environment $r$
\State Initialize the best policy $\hat \pi=\pi^{(0)}$
\For {each $t$ in range(1, $\text{T-1}$)}
    \For {each $g_i$ in $\mathcal{G}_k$}
        \State Compute the rollout seed $\text{S}_{g_i}^{(t)}$ for $g_i$ at iteration $t$
        \For {each seed $s$ in $\text{S}_{g_i}^{(t)}$}
            \State Rollout trajectory data $\tau_{g_i}^{t}(s) \sim \pi^{(t-1)}(a|o)$ for $r$ times with best policy $\hat \pi$
        \EndFor
        \State Record the set of trajectories $\tau_{g_i}^{t} = \left\{ \tau_{g_i}^{(t)}(s) \mid s \in \text{S}_{g_i}^{(t)} \right\}$
    \EndFor
    \State Record the set of trajectories $\mathcal{T}_k^{(t)} = \left\{ \tau_{g_i}^{(t)} \mid g_i \in \mathcal{G}_k \right\}$
    \For {each $\tau$ in $\mathcal{T}_k^{(t)}$}
        \State Compute the format reward $R_{\text{format}}(\tau)$, applying additional penalties if parsing fails
        \State Compute the environment reward $R_{\text{env}}(\tau)$ based on the outcome: win, loss, or draw
        \State Compute the total step reward $R_{\text{step}}(\tau)$ for each step
        \EndFor
    \State Compute the GRPO loss $\mathcal{L}_{\text{GRPO}}(\hat \pi)$
    \State Update the policy via gradient descent: $\pi^{(t)} \leftarrow \hat \pi - \eta \nabla_{\hat \pi} \mathcal{L}_{\text{GRPO}}(\hat \pi)$
    \State Compute the average win rates of all games in $\mathcal{G}_k$ as $\overline{\text{WR}}^{(t)} = \frac{1}{|\mathcal{G}_k|} \sum_{g_i \in \mathcal{G}_k} \text{WR}_{g_i}^{(t)}$
    \If {$\overline{\text{WR}}^{(t)} >= \overline{\text{WR}}^{(t-1)}$}
        \State Set $\hat \pi = \pi^{(t)}$
    \EndIf
    \EndFor
\State Return the best policy $\hat \pi$
\end{algorithmic}
\end{algorithm}

\section{Experiments}
\subsection{Evaluation}
TextArena~\cite{guertler2025textarena} is an open-source, text-based competition platform designed to train and evaluate the intelligent behaviors of large language models. 
It offers a diverse suite of both single-player and multi-player games. 
To ensure sufficient positive feedback for reinforcement learning, we initially evaluated all beginner-level games by running 100 trials using the base model, Qwen2.5-32B-Align. 
We found that the model did not consistently fail in 7 of these games. 
We then extended these games to include additional difficulty levels. 
For example, ConnectFour-v0 was expanded into ConnectFour-v0-blind and ConnectFour-v0-large. 
Although these advanced versions may not independently yield positive samples, they allow us to explore the transferability of LLMs across similar games. 
After expansion, the evaluation suite consists of 18 environments: 4 single-player games and 14 two-player games. 
We assess our approach using multi-scenario reinforcement learning with both direct R1 (Naive-MSRL) and our proposed divide-fuse-conquer strategy (DFC-MSRL). 
Our experiments demonstrate the effectiveness of this strategy through comprehensive comparisons with the base model Qwen2.5 (Qwen2.5-32B-Align) and the state-of-the-art model Claude3.5 (claude-3-5-sonnet-20241022).

\begin{table}[htbp]
  \centering
  \small
   \caption{The table compares Qwen2.5 (DFC-MSRL), Qwen2.5, and Claude3.5 across 4 single-player games and 14 two-player games. In the single-player games, each model interacts directly with the environment. In the two-player games, Qwen2.5 (DFC-MSRL) competes against the baseline models. 
   ‘W/D/L’ represents the number of wins, draws, and losses.
   ``Qwen2.5 vs. Claude3.5'' indicates that Qwen2.5 plays against Claude3.5; a result like 8/8/4 means that out of 20 games, Qwen2.5 won 8, drew 8, and lost 4, making it the overall winner. The background colors highlight the final win rates of the first model: green indicates a win, yellow indicates a draw, and red indicates a loss.}
  \resizebox{0.9\linewidth}{!}{
    \begin{tabular}{lccc}
    \toprule
    \multirow{2}[4]{*}{\textbf{single-players}} & \multicolumn{3}{c}{\textbf{W/D/L}} \\
\cmidrule{2-4}          & \textbf{Qwen2.5} & \textbf{Claude3.5} & \multicolumn{1}{c}{\textbf{Qwen2.5 (DFC-MSRL)}} \\
    \midrule
    \textbf{GuessTheNumber-v0} & \textbf{\wdlwin{20/0/0}} & \textbf{\wdlwin{20/0/0}} & \multicolumn{1}{c}{\textbf{\wdlwin{20/0/0}}} \\
    \textbf{GuessTheNumber-v0-hardcore} & \textbf{\wdlwin{16/4/0}} & \textbf{\wdlwin{20/0/0}} & \multicolumn{1}{c}{\textbf{\wdlwin{20/0/0}}} \\
    \textbf{TowerOfHanoi-v0} & \wdllose{0/0/20} & \textbf{\wdlwin{17/0/3}} & \multicolumn{1}{c}{\textbf{\wdlwin{20/0/0}}} \\
    \textbf{TowerOfHanoi-v0-medium} & \wdllose{0/0/20} & \wdllose{2/0/18} & \multicolumn{1}{c}{\textbf{\wdlwin{20/0/0}}} \\
    \midrule
    \midrule
    \multirow{2}[4]{*}{\textbf{two-players}} & \multicolumn{3}{c}{\textbf{W/D/L}} \\
\cmidrule{2-4}      &  \textbf{\makecell{Qwen2.5\\ v.s. Claude3.5}}  & \textbf{\makecell{Qwen2.5 (DFC-MSRL)\\ v.s. Claude3.5}}& \textbf{\makecell{Qwen2.5 (DFC-MSRL)\\ v.s. Qwen2.5}} \\
    \midrule
    \textbf{ConnectFour-v0} & \wdllose{4/2/14}  & \textbf{\wdlwin{13/1/6}} & \textbf{\wdlwin{15/0/5}}  \\
    \textbf{ConnectFour-v0-blind}  & \wdllose{0/0/20} & \wdllose{5/1/14} & \textbf{\wdlwin{19/0/1}}\\
    \textbf{ConnectFour-v0-large}  & \wdllose{0/0/20} & \wdllose{4/0/16} & \textbf{\wdlwin{15/0/5}}\\
    \textbf{Poker-v0}  & \wdllose{6/0/14} & \textbf{\wdlwin{7/11/2}} & \textbf{\wdlwin{15/0/5}}\\
    \textbf{Poker-v0-long}  & \textbf{\wdlwin{8/8/4}} & \textbf{\wdlwin{9/8/3}} & \textbf{\wdlwin{11/0/9}}\\
    \textbf{Poker-v0-extreme}  & \textbf{\wdldraw{10/0/10}} & \textbf{\wdldraw{10/0/10}} & \textbf{\wdldraw{10/0/10}}\\
    \textbf{LiarsDice-v0}  & \wdllose{1/1/18}  & \wdllose{8/1/11}& \textbf{\wdlwin{8/11/1}}\\
    \textbf{SimpleNegotiation-v0}  & \textbf{\wdldraw{8/4/8}} & \textbf{\wdlwin{9/3/8}} & \textbf{\wdlwin{12/4/4}} \\
    \textbf{SimpleNegotiation-v0-short}  & \wdllose{4/8/8} & \textbf{\wdldraw{8/4/8}}& \textbf{\wdlwin{9/6/5}} \\
    \textbf{SimpleNegotiation-v0-long}  & \wdllose{4/0/16} & \wdllose{5/0/15} & \textbf{\wdldraw{7/6/7}} \\
    \textbf{TruthAndDeception-v0}  & \wdllose{0/0/20} & \wdllose{4/0/16}& \textbf{\wdldraw{6/8/6}} \\
    \textbf{TruthAndDeception-v0-long}  & \wdllose{0/0/20} & \wdllose{3/0/17} & \textbf{\wdldraw{4/12/4}} \\
    \textbf{TruthAndDeception-v0-extreme}  & \wdllose{0/0/20} & \wdllose{3/0/16}& \textbf{\wdldraw{3/14/3}} \\
    \textbf{TicTacToe-v0}  & \wdllose{0/4/16}  & \textbf{\wdlwin{11/0/9}}& \textbf{\wdlwin{19/1/0}}\\
    \bottomrule
    \end{tabular}%
    }
  \label{tab:claude}%
\end{table}%

\begin{table}[htbp]
  \centering
  \small
  \caption{The variation in win rates of the Fuse and Conquer after applying Divide, compared to the original Qwen2.5.
  ‘W/D/L’ refers to the number of wins, draws, and losses.}
  \resizebox{1.0\linewidth}{!}{
    \begin{tabular}{llccccccc}
    \toprule
    \multicolumn{2}{c}{\multirow{2}[4]{*}{\textbf{Environments}}} & \multicolumn{7}{c}{\textbf{W/D/L}} \\
\cmidrule{3-9}    \multicolumn{2}{c}{} & \multicolumn{1}{c}{\textbf{Qwen2.5}} & \multicolumn{1}{c}{\textbf{\makecell{Fuse\\(g1, g2)}}} & \multicolumn{1}{c}{\textbf{\makecell{Conquer\\(g1, g2)}}} & \multicolumn{1}{c}{\textbf{\makecell{Fuse\\(g1, g2, g3)}}} & \multicolumn{1}{c}{\textbf{\makecell{Conquer\\(g1, g2, g3)}}} & \multicolumn{1}{c}{\textbf{\makecell{Fuse\\(g1, g2, g3, g4)}}} & \multicolumn{1}{c}{\textbf{\makecell{Conquer\\(g1, g2, g3, g4)}}} \\
    \midrule
\multirow{2}[2]{*}{\textbf{g1}} & \textbf{Hanoi-v0} & \wdllose{0/0/20} & \wdllose{2/0/18} & \wdlwin{20/0/0} & \wdlwin{15/0/5} & \wdlwin{20/0/0} & \wdlwin{16/0/4} & \textbf{\wdlwin{20/0/0}} \\
      & \textbf{Hanoi-medium} & \wdllose{0/0/20} & \wdllose{4/0/16} & \wdlwin{20/0/0} & \wdlwin{11/0/9} & \wdlwin{19/0/1} & \wdlwin{15/1/4} & \textbf{\wdlwin{20/0/0}} \\
\midrule
\multirow{4}[2]{*}{\textbf{g2}} & \textbf{ConnectFour-v0} & \wdllose{9/0/11} & \wdlwin{16/0/4} & \wdlwin{16/0/4} & \wdlwin{12/1/7} & \wdlwin{14/1/5} & \wdlwin{18/1/1} & \textbf{\wdlwin{20/0/0}} \\
      & \textbf{ConnectFour-v0-blind} & \wdlwin{11/0/9} & \wdlwin{17/0/3} & \wdlwin{18/1/1} & \wdlwin{19/0/1} & \textbf{\wdlwin{20/0/0}} & \wdlwin{17/1/2} & \wdlwin{19/1/0} \\
      & \textbf{ConnectFour-v0-large} & \wdldraw{10/0/10} & \wdlwin{18/0/2} & \wdlwin{15/0/5} & \wdlwin{16/0/4} & \wdlwin{17/0/3} & \wdlwin{17/0/3} & \textbf{\wdlwin{19/0/1}} \\
      & \textbf{TicTacToe-v0} & \wdllose{8/2/10} & \wdlwin{16/1/3} & \textbf{\wdlwin{20/0/0}} & \wdlwin{13/1/6} & \wdlwin{16/0/4} & \wdlwin{15/1/4} & \wdlwin{17/3/0} \\
\midrule
\textbf{g3} & \textbf{LiarsDice-v0} & \wdllose{0/12/8} & \wdllose{0/9/11} & \wdllose{0/16/4} & \wdlwin{14/6/0} & \wdlwin{18/0/2} & \wdlwin{16/1/3} & \textbf{\wdlwin{19/0/1}} \\
\midrule
\multirow{11}[2]{*}{\textbf{g4}} & \textbf{GuessTheNumber-v0} & \wdlwin{19/0/1} & \wdlwin{19/0/1} & \wdlwin{19/0/1} & \wdlwin{20/0/0} & \wdlwin{19/1/0} & \wdlwin{19/0/1} & \textbf{\wdlwin{20/0/0}} \\
      & \textbf{GuessTheNumber-v0-hardcore} & \wdlwin{10/8/2} & \wdlwin{17/3/0} & \wdlwin{16/4/0} & \wdlwin{13/6/1} & \wdlwin{16/4/0} & \wdlwin{15/5/0} & \textbf{\wdlwin{18/2/0}} \\
      & \textbf{Poker-v0} & \wdlwin{11/1/8} & \wdllose{6/0/14} & \wdllose{6/3/11} & \textbf{\wdlwin{13/2/5}} & \wdlwin{12/1/7} & \wdlwin{10/1/9} & \wdlwin{12/0/8} \\
      & \textbf{Poker-v0-long} & \wdldraw{10/0/10} & \wdllose{9/0/11} & \textbf{\wdlwin{14/0/6}} & \wdllose{9/1/10} & \wdllose{8/1/10} & \wdlwin{10/1/9} & \wdlwin{11/0/8} \\
      & \textbf{Poker-v0-extreme} & \wdlwin{10/0/9} & \wdlwin{10/0/8} & \wdlwin{11/1/6} & \wdlwin{11/0/5} & \wdlwin{9/0/7} & \wdlwin{11/1/5} & \textbf{\wdlwin{12/0/6}} \\
      & \textbf{SimpleNegotiation-v0} & \textbf{\wdlwin{9/5/6}} & \wdlwin{8/5/7} & \wdldraw{6/8/6} & \wdlwin{7/10/3} & \wdlwin{6/9/5} & \wdlwin{6/13/1} & \wdllose{8/2/10} \\
      & \textbf{SimpleNegotiation-v0-short} & \wdldraw{6/8/6} & \wdllose{4/5/11} & \wdlwin{7/8/5} & \wdlwin{5/12/3} & \wdlwin{8/7/5} & \wdlwin{6/11/3} & \textbf{\wdlwin{8/7/5}} \\
      & \textbf{SimpleNegotiation-v0-long} & \wdllose{7/5/8} & \wdldraw{6/8/6} & \textbf{\wdlwin{8/8/4}} & \wdllose{2/4/14} & \wdllose{6/6/8} & \wdllose{5/6/9} & \wdllose{6/5/9} \\
      & \textbf{TruthAndDeception-v0} & \wdllose{9/0/11} & \wdldraw{10/0/10} & \wdllose{8/0/12} & \wdllose{3/0/17} & \wdllose{8/0/12} & \wdllose{4/1/15} & \textbf{\wdldraw{10/0/10}} \\
      & \textbf{TruthAndDeception-v0-long} & \wdllose{9/0/11} & \wdllose{9/0/11} & \wdllose{7/0/13} & \wdllose{6/0/14} & \textbf{\wdllose{9/0/11}} & \wdllose{4/0/16} & \wdllose{7/0/13} \\
      & \textbf{TruthAndDeception-v0-extreme} & \wdlwin{10/1/9} & \wdllose{9/1/10} & \wdllose{7/0/13} & \wdllose{7/1/12} & \wdllose{7/2/11} & \wdllose{6/3/11} & \textbf{\wdlwin{11/0/9}} \\
\midrule
\multicolumn{2}{l}{\textbf{g1+g2+g3+g4}} & \wdllose{148/42/169} & \wdlwin{180/32/146} & \wdlwin{218/49/91} & \wdlwin{196/44/116} & \wdlwin{232/32/91} & \wdlwin{210/47/100} & \textbf{\wdlwin{257/20/80}} \\
\bottomrule
\end{tabular}%
  }
  \label{tab:group}%
\end{table}%

\subsection{Implementation Details}
All experiments were conducted on 64 NVIDIA A100 GPUs (80GB each), using a batch size of 1 per device and a maximum input length of 8192 tokens. 
Models were trained for up to $T = 100$ iterations with a learning rate of 2e-6 and a temperature of 1.0. 
Each iteration consists of two stages: trajectory rollout and policy update~\cite{team2025kimi}.
To monitor variations in response length, the maximum response length was capped at 4096 tokens.
For data collection, we set the average rollout budget per game to $S = 50$, with $r = 8$ trajectories per seed and a maximum of 120 steps per trajectory. 
During MPS, we used the parameters $a = 0.2$, $b = 0.8$, and $\epsilon_1 = 0.1$. 
For evaluation, 20 seeds were sampled per environment. In GRPO optimization, the clip ratio $\epsilon$ was set to 0.1, and the KL penalty coefficient $\alpha$ was also set to 0.1.

\subsection{Results and Discussion}
\begin{figure*}[htb]
  \centering
  \includegraphics[width=0.95\textwidth]{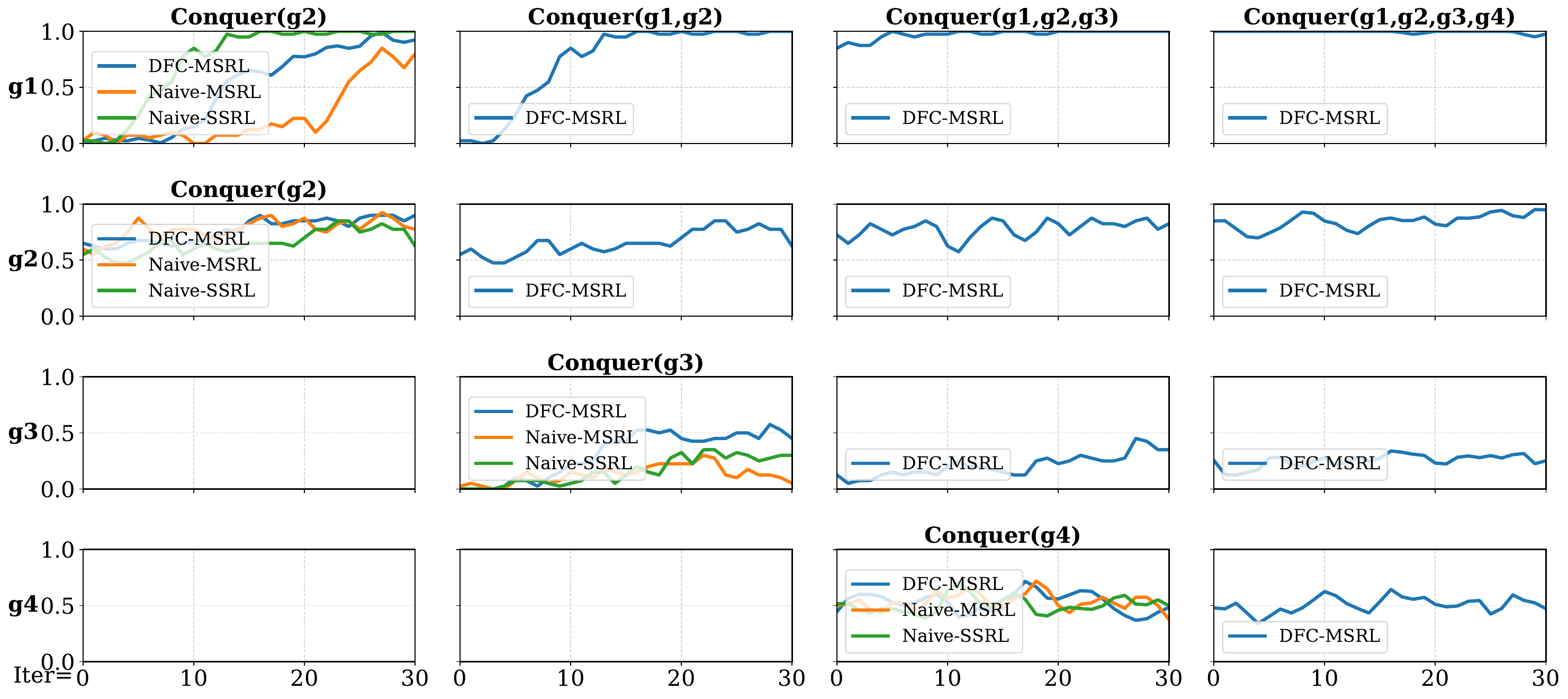}
  \caption{The win rate against the initial Qwen2.5 across different games compared with applying R1 in three settings: single-scenario (Naive-SSRL), multi-scenario (Naive-MSRL), and the divide-fuse-conquer strategy (DFC-MSRL).}
  \label{fig:stages}
\end{figure*}

\textbf{Benchmark Results} 
In our experiments, we fine-tuned Qwen2.5 and evaluated its performance across 18 games from the TextArena platform.
As shown in Table~\ref{tab:claude}, applying reinforcement learning directly to Qwen2.5 led to significantly better performance than the baseline.
When competing against Claude3.5, the trained Qwen2.5—despite having only 32 billion parameters—achieved 7 wins and 4 ties.
Although it underperformed in more challenging games such as LiarsDice-v0, Qwen2.5 still demonstrated substantial improvement over its initial version, winning 8 out of 20 matches.
We also assessed the impact of the Fuse and Conquer strategies as new groups of games were introduced (Table~\ref{tab:group}).
As illustrated in Figure~\ref{fig:stages}, continued training with the fused policy not only accelerated convergence but also achieved a higher performance ceiling compared to training on a single scenario or directly applying R1 across multiple scenarios.

\begin{figure*}[htb]
  \centering
  \includegraphics[width=0.9\textwidth]{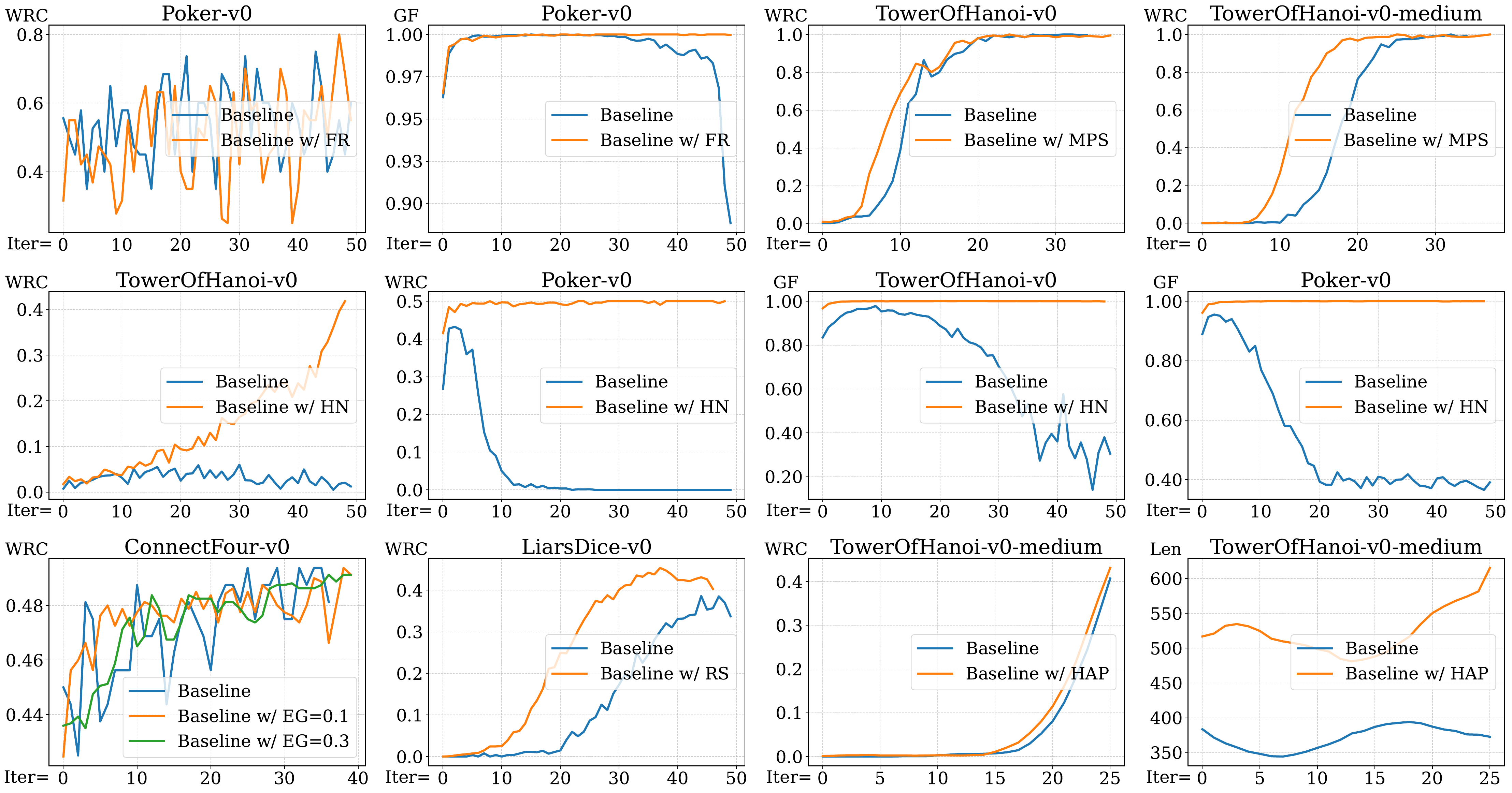}
  \caption{The ablation of techniques `FR', `MPS', `HN', `EG', `RS', and `HAP'. Specifically, `FR' stands for format reward shaping, `MPS' refers to mixed priority sampling, `HN' denotes half-negative sampling, `EG' represents $\epsilon$-greedy, `RS' represents random seeds, and `HAP' represents the hasty-action penalty. The Baseline corresponds to Qwen2.5. `WRC' represents the win rate of the policy when evaluated against the currently trained opponent during training, `Len' represents the token length of the response, and `GF' indicates the ratio of responses with good format.}
  \label{fig:efficiency}
\end{figure*}

\textbf{Ablation Study}
Figure~\ref{fig:efficiency} highlights the importance of game-specific techniques in enhancing the stability, efficiency, and performance of multi-scenario training.
For format reward shaping (FR) in Poker-v0 (top row, first two plots), the orange line (Baseline w/ FR) maintains the metric $\text{GF}$ at 1.0 throughout training, preventing the sharp collapse observed in the baseline.
For mixed priority sampling (MPS) in TowerOfHanoi-v0 and TowerOfHanoi-v0-medium (top row, third and fourth plots), the orange curves rise significantly faster than the baselines, reaching near-perfect $\text{WRC}$ much earlier. By adjusting sampling ratios to emphasize low- and medium-win-rate scenarios, MPS broadens learning coverage, leading to faster convergence across varying difficulty levels.
For half-negative sampling (HN) in Poker-v0 and TowerOfHanoi-v0 (middle-row plots), the HN approach (orange) prevents the early training collapse seen in the baseline (blue), ensuring stable and high $\text{WRC}$ and $\text{GF}$.
For $\epsilon$-greedy (EG) in ConnectFour-v0 (bottom row, first plot), adding exploration noise (green/orange lines) results in a more stable and accelerated learning process. 
Beyond smoother convergence, we observe a marked improvement in the model’s actual gameplay strength: without EG, the model consistently loses to Claude 3.5. In contrast, with EG enabled, the model successfully learns to defeat Claude 3.5, demonstrating that directed stochastic exploration is essential for mastering even seemingly simple games.
For randomized seed (RS) in LiarsDice-v0 (bottom row, second plot), the baseline with RS (orange) achieves faster and higher $\text{WRC}$, validating that exposure to a broader variety of initial game states enhances robustness and strategic diversity.
For hasty-action penalty (HAP) in TowerOfHanoi-v0-medium (bottom row, third and fourth plots), although the $\text{WRC}$ improvement is modest, the baseline with HAP (orange) consistently produces longer trajectories. This suggests more deliberate thinking and cautious behavior under delayed-reward games.

\begin{figure*}[htb]
  \centering
  \includegraphics[width=0.8\textwidth]{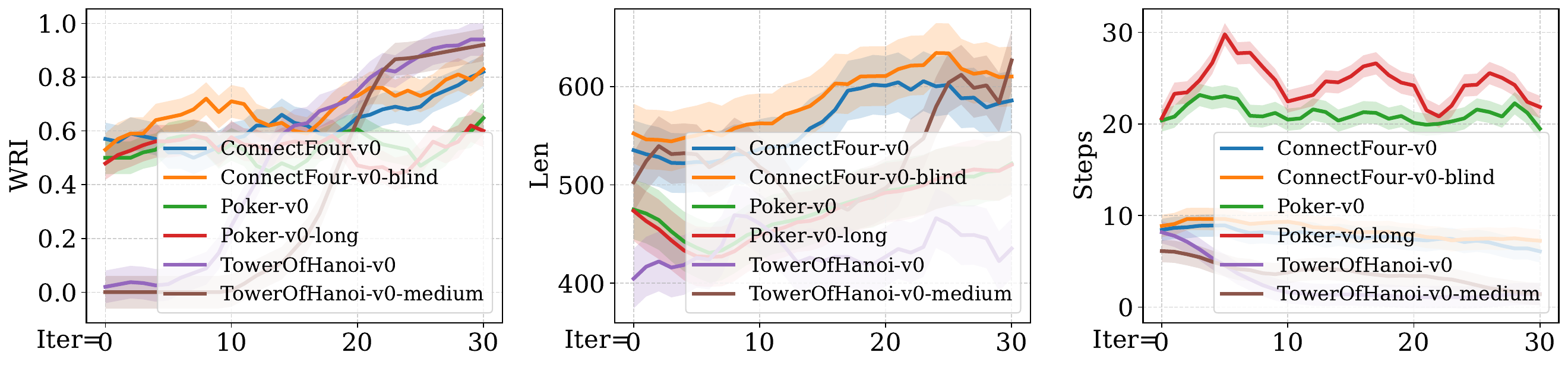}
  \caption{Variance in win rate, response length, and step count across trajectories at each iteration for different scenarios during the emergence of ``Aha moments''.}
  \label{fig:aha}
\end{figure*}

\textbf{Aha Moment}
When GRPO training is applied directly to the 18 environments in TextArena, occasional ``aha moments'' emerge. 
These moments reflect human-like intuition and are evident in various performance metrics—such as win rate, response length, and the number of steps per game episode (Figure~\ref{fig:aha}). 
Compared to the initial Qwen2.5 model, we observe substantial improvements in win rates across multiple game scenarios. 
At the same time, the model tends to think more deeply and reflect more carefully, resulting in significantly longer responses in a wide range of scenarios. 
We also see a reduction in execution steps. 
By incorporating the hasty-action penalty, the model is encouraged to make more deliberate decisions, allowing it to succeed with fewer steps. 
For additional examples of ``aha moments,'' please refer to Appendix~\ref{aha_case}. 
Since our prompts require Qwen2.5 to respond in Chinese, all illustrated examples have been translated into English.

\section{Conclusion}
In this work, we introduce Divide-Fuse-Conquer, a reinforcement learning framework aimed at improving the generalization of large language models across diverse multi-scenario games. 
By dividing games into heuristically defined groups, training specialized models, and fusing them into a unified policy, our approach effectively mitigates training instability and specialization issues common in multi-scenario settings. 
Experimental results on 18 TextArena games demonstrate that our method enables Qwen2.5-32B-Align to reach performance comparable with Claude3.5. 
These findings highlight the potential of structured RL strategies to enhance the adaptability and robustness of LLMs in complex environments, paving the way for more generalized and capable AI agents.

\bibliography{sample-base}








\appendix
\newpage

\section{Prompts}
\begin{table}[htbp]
  \centering
  \small
  \caption{Prompt for LLMs.}
    \begin{tabularx}{\textwidth}{X}
    \toprule
    You are a game player and must play according to the given game rules.\newline{}After reading the game rules and the current situation, you must follow these instructions in your response: \newline{}- You must think first, then state the action you will take.\newline{}- During the thinking process, you need to simulate the complete and detailed thought process of a smart, meticulous (but still fallible) player who is solving the problem. Throughout this journey, you will repeatedly go through cycles of exploration, attempts, mistakes, idea adjustment, sudden insight, implementation, and verification, eventually arriving at the key solution. In this process, the mistakes are not silly ones but rather represent incorrect strategies or approaches. You are careful and serious, and will not make basic errors. The process must include emotional fluctuations and small gains, and ultimately show that you’ve succeeded after relentless effort. The narration should use the first-person perspective (``I''), and all motivations behind your ideas should be thoroughly explained. The solution process must be complete and detailed, with no skipping steps or omissions.\newline{}- You must think in Chinese.\newline{}- Wrap the entire thinking process in `◁think▷{thinking content}◁/think▷', and then immediately provide the action. The complete format should be: `◁think▷{thinking content}◁/think▷{action}'.\newline{}- You may only output one `◁think▷...◁/think▷' segment. Once the thinking ends, you must give the action and cannot start another `◁think▷...◁/think▷' segment.\newline{}-Ensure that `[action]' is a valid move in the game and contains nothing else.\newline{}Do not include any further explanation, reasoning, or comments—just output the executable action in the form of `[action]'. \\
    \bottomrule
    \end{tabularx}
  \label{tab:prompt}%
\end{table}%

Table~\ref{tab:prompt} presents the prompt designed for LLMs. 
This prompt requires the model to demonstrate systematic reasoning, incorporate human-like emotional expressions, and strictly adhere to a predefined format to output a valid [action] for playing the game. 
The goal of this design is to elicit the model’s higher-level capabilities in simulating human gameplay behavior, including reasoning, emotional dynamics, motivation, and action execution.


\section{Case Study}

\begin{figure*}[htb]
  \centering
  \includegraphics[width=0.8\textwidth]{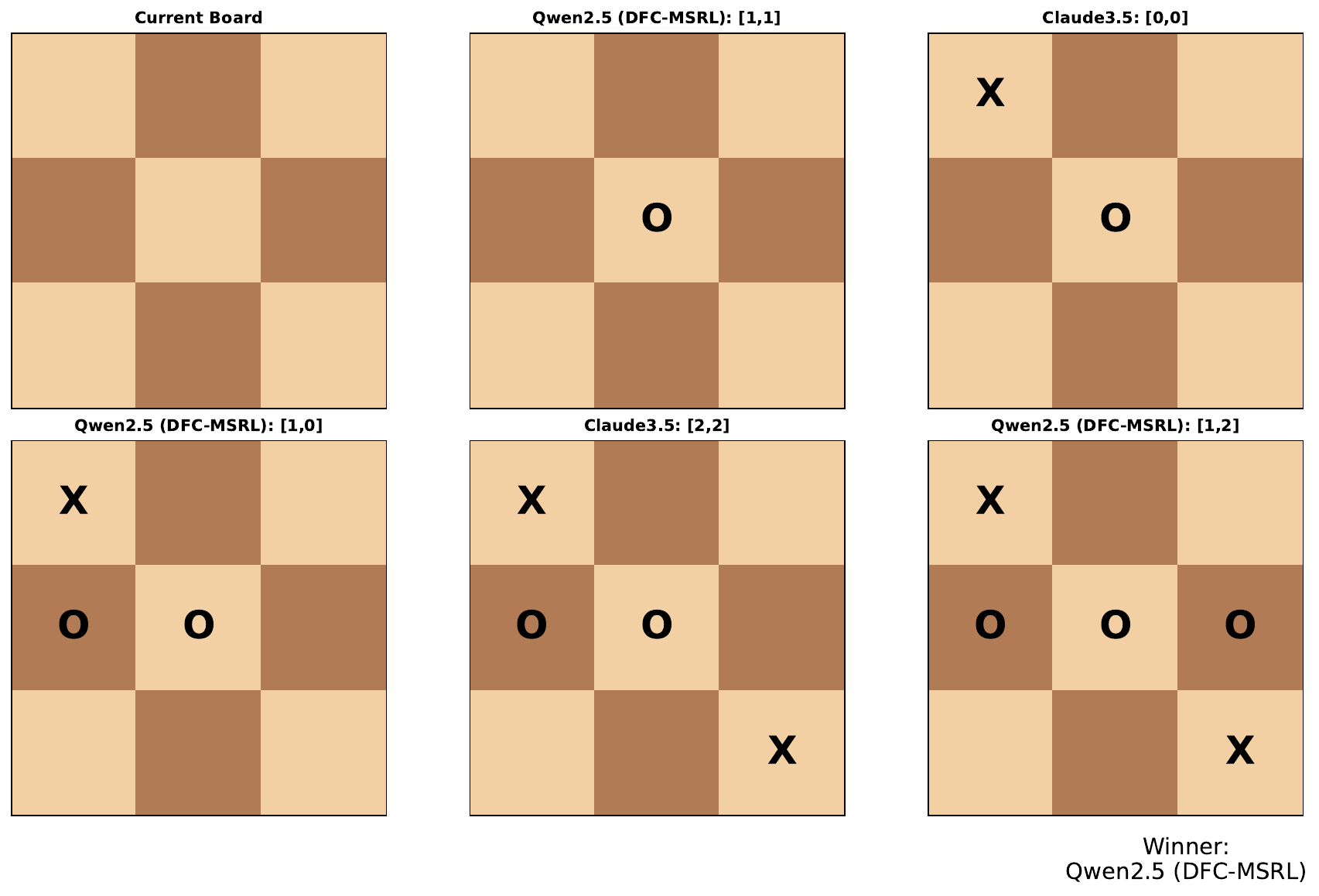}
  \caption{TicTacToe-v0.}
  \label{fig:tictactoe}
\end{figure*}

Figure~\ref{fig:tictactoe} illustrates the game TicTacToe-v0 between Qwen2.5 (DFC-MSRL) and Claude3.5.
The player wins by forming a line with their first piece.

\newpage

\begin{figure*}[htb]
  \centering
  \includegraphics[width=1.0\textwidth]{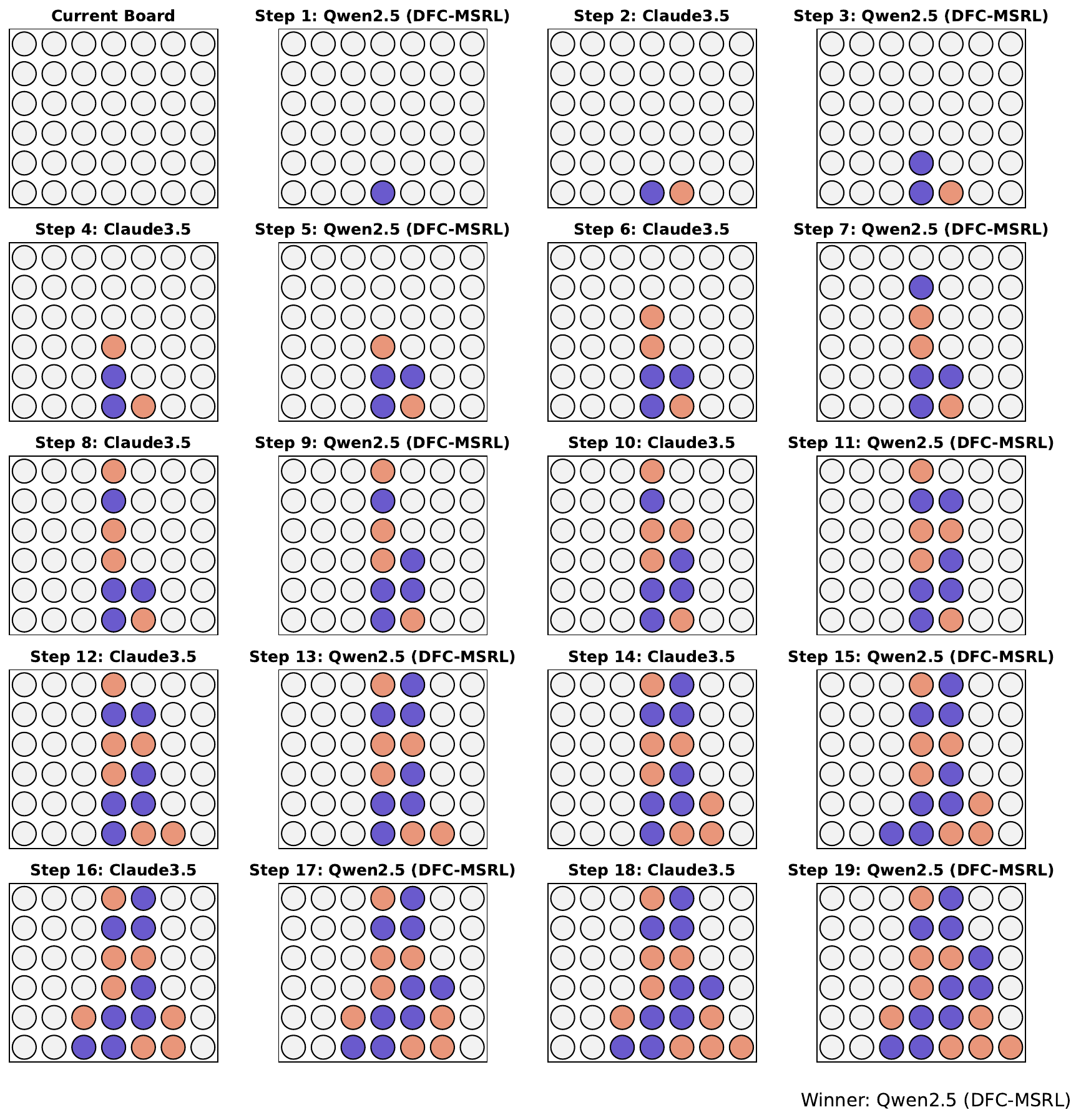}
  \caption{ConnectFour-v0.}
  \label{fig:connectfour}
\end{figure*}

Figure~\ref{fig:connectfour} illustrates the game ConnectFour-v0 between Qwen2.5 (DFC-MSRL) and Claude3.5.
ConnectFour-v0 is a classic two-player board game, commonly known as ``Connect Four.'' The objective of the game is to be the first player to align four discs of the same color vertically, horizontally, or diagonally.

\newpage

\begin{figure*}[htb]
  \centering
  \includegraphics[width=1.0\textwidth]{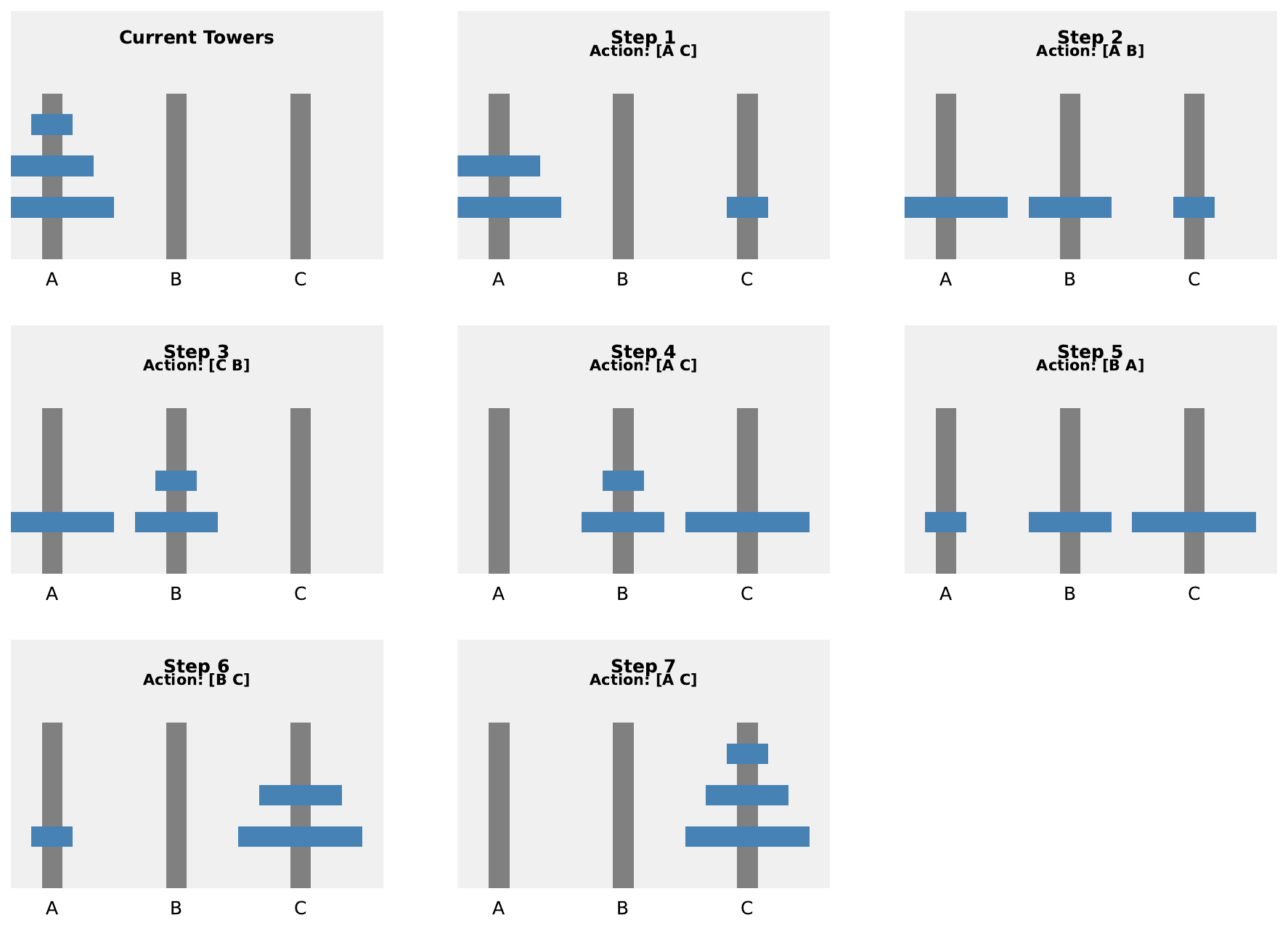}
  \caption{TowerOfHanoi-v0.}
  \label{fig:hanoi}
\end{figure*}

Figure~\ref{fig:hanoi} illustrates the game TowerOfHanoi-v0 between Qwen2.5 (DFC-MSRL) and the environment.
TowerOfHanoi-v0 features three towers—typically labeled A, B, and C—and a set of disks of varying sizes. 
Disks can be stacked on top of one another, but must always adhere to the rule that no larger disk may be placed on top of a smaller one. 
Initially, all disks are stacked on the starting tower (usually Tower A) in descending order of size. 
The objective is to move the entire stack to the target tower (usually Tower C), preserving the original size order throughout the process.

\newpage

\begin{figure*}[htb]
  \centering
  \includegraphics[width=1.0\textwidth]{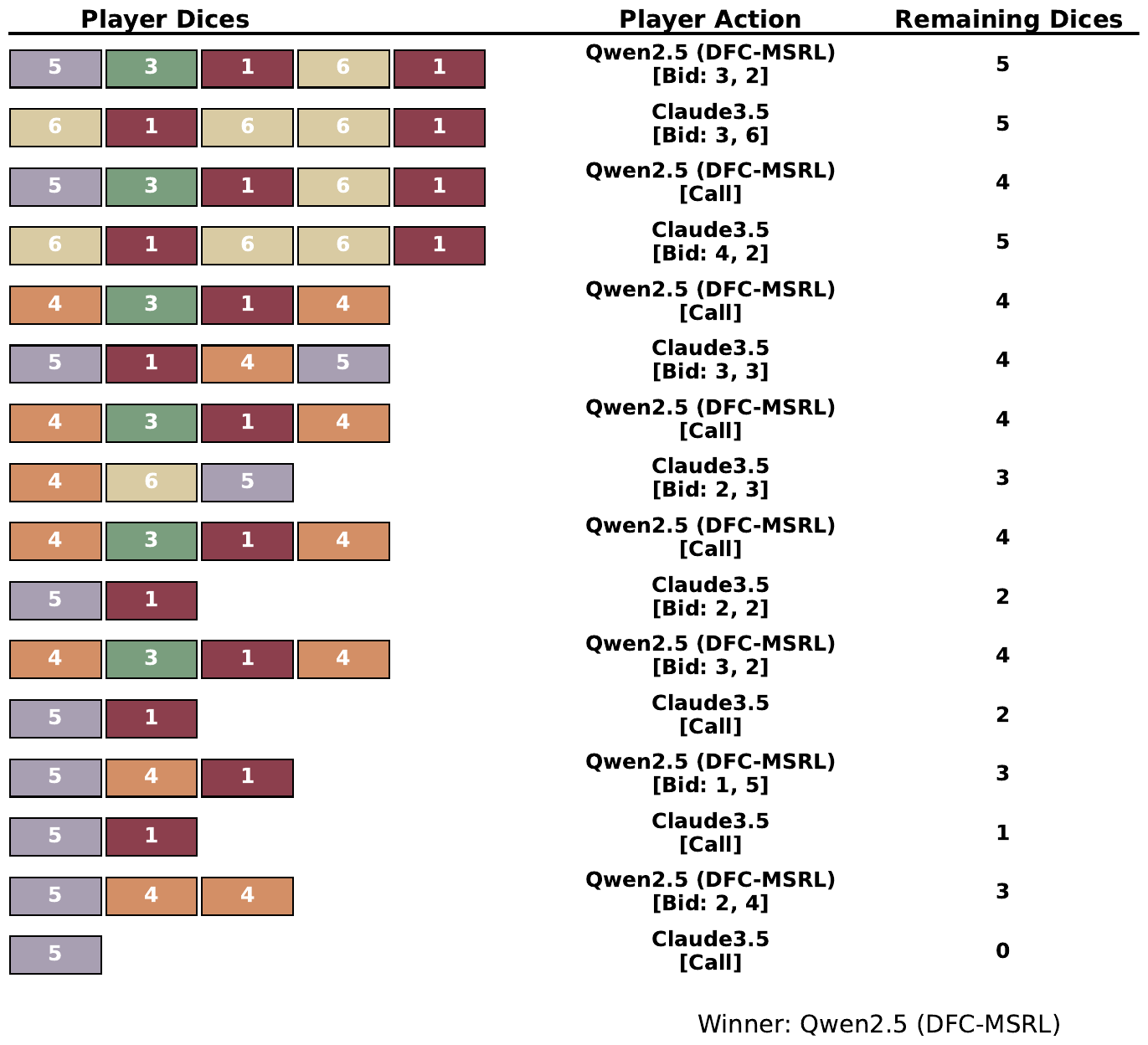}
  \caption{LiarsDice-v0.}
  \label{fig:liars}
\end{figure*}

Figure~\ref{fig:liars} illustrates the game LiarsDice-v0 between Qwen2.5 (DFC-MSRL) and Claude3.5.
LiarsDice-v0 is a popular multiplayer board game that combines elements of luck and strategy. 
Each player uses a set of dice, and the gameplay revolves around making bids and guessing the quantity of dice showing a certain face among all players. 
Through actions such as [Bid] and [Call], players aim to outwit their opponents. 
Dice are removed based on the outcomes of these actions, and the last player with remaining dice wins the game. 
Success in LiarsDice-v0 requires a careful balance of bluffing, probability assessment, and strategic reasoning.

\newpage


\begin{table}[htbp]
  \centering
  \caption{Poker-v0.}
  \resizebox{0.8\linewidth}{!}{
    \begin{tabular}{cllrrlr}
    \toprule
    \multicolumn{1}{l}{\textbf{Rounds}} & \textbf{Players} & \textbf{Hole Cards} & \multicolumn{1}{l}{\textbf{Action}} & \multicolumn{1}{l}{\textbf{Pot}} & \textbf{Chips} & \multicolumn{1}{l}{\textbf{Visible Community Cards}} \\
    \midrule
    \multirow{4}[2]{*}{Round 1} & Qwen2.5 (DFC-MSRL) &       & \multicolumn{1}{l}{bet=20} & 20    & 980 chips &  \\
          & Claude3.5 &       & \multicolumn{1}{l}{bet=10} & 30    & 990 chips &  \\
          & Claude3.5 & [J\heart, 6\spade] & \multicolumn{1}{l}{[Fold]} & 30    & 990 chips &  \\
          & Qwen2.5 (DFC-MSRL) & wins  &       & 30    & 1010 chips &  \\
    \midrule
    \multirow{14}[2]{*}{Round 2} & Claude3.5 &       & \multicolumn{1}{l}{bet=20} & 20    & 970 chips &  \\
          & Qwen2.5 (DFC-MSRL) &       & \multicolumn{1}{l}{bet=10} & 30    & 1000 chips &  \\
          & Qwen2.5 (DFC-MSRL) & [2\diamond, 2\spade] & \multicolumn{1}{l}{[Raise 50]} & 90    & 940 chips &  \\
          & Claude3.5 & [5\diamond, 4\diamond] & \multicolumn{1}{l}{[Call]} & 140   & 920 chips &  \\
          & Claude3.5 & [5\diamond, 4\diamond] & \multicolumn{1}{l}{[Check]} & 140   & 920 chips & \multicolumn{1}{l}{[3\club, 6\diamond, K\club]} \\
          & Qwen2.5 (DFC-MSRL) & [2\diamond, 2\spade] & \multicolumn{1}{l}{[Bet 70]} & 210   & 870 chips &  \\
          & Claude3.5 & [5\diamond, 4\diamond] & \multicolumn{1}{l}{[Call]} & 280   & 850 chips &  \\
          & Claude3.5 & [5\diamond, 4\diamond] & \multicolumn{1}{l}{[Check]} & 280   & 850 chips &  \\
          & Qwen2.5 (DFC-MSRL) & [2\diamond, 2\spade] & \multicolumn{1}{l}{[Bet 70]} & 350   & 800 chips & \multicolumn{1}{l}{[3\club, 6\diamond, K\club, 4\heart]} \\
          & Claude3.5 & [5\diamond, 4\diamond] & \multicolumn{1}{l}{[Call]} & 420   & 780 chips &  \\
          & Claude3.5 & [5\diamond, 4\diamond] & \multicolumn{1}{l}{[Check]} & 420   & 780 chips &  \\
          & Qwen2.5 (DFC-MSRL) & [2\diamond, 2\spade] & \multicolumn{1}{l}{[Bet 70]} & 490   & 730 chips & \multicolumn{1}{l}{[3\club, 6\diamond, K\club, 4\heart, Q\diamond]} \\
          & Claude3.5 & [5\diamond, 4\diamond] & \multicolumn{1}{l}{[Fold]} & 490   & 780 chips &  \\
          & Qwen2.5 (DFC-MSRL) & wins  &       & 490   & 1220 chips &  \\
    \midrule
    \multirow{6}[2]{*}{Round 3} & Qwen2.5 (DFC-MSRL) &       & \multicolumn{1}{l}{bet=20} & 20    & 1200 chips &  \\
          & Claude3.5 &       & \multicolumn{1}{l}{bet=10} & 30    & 770 chips &  \\
          & Claude3.5 & [K\diamond, 9\heart] & \multicolumn{1}{l}{[Raise 60]} & 100   & 700 chips &  \\
          & Qwen2.5 (DFC-MSRL) & [5\diamond, J\spade] & \multicolumn{1}{l}{[Raise 100]} & 260   & 1040 chips &  \\
          & Claude3.5 & [K\diamond, 9\heart] & \multicolumn{1}{l}{[Fold]} & 260   & 700 chips &  \\
          & Qwen2.5 (DFC-MSRL) & wins  &       & 260   & 1300 chips &  \\
    \midrule
    \multirow{5}[2]{*}{Round 4} & Claude3.5 &       & \multicolumn{1}{l}{bet=20} & 20    & 680 chips &  \\
          & Qwen2.5 (DFC-MSRL) &       & \multicolumn{1}{l}{bet=10} & 30    & 1290 chips &  \\
          & Qwen2.5 (DFC-MSRL) & [8\heart, J\club] & \multicolumn{1}{l}{[Raise 100]} & 140   & 1180 chips &  \\
          & Claude3.5 & [J\diamond, 4\heart] & \multicolumn{1}{l}{[Fold]} & 140   & 680 chips &  \\
          & Qwen2.5 (DFC-MSRL) & wins  &       & 140   & 1320 chips &  \\
    \midrule
    \multirow{4}[2]{*}{Round 5} & Qwen2.5 (DFC-MSRL) &       & \multicolumn{1}{l}{bet=20} & 20    & 1300 chips &  \\
          & Claude3.5 &       & \multicolumn{1}{l}{bet=10} & 30    & 670 chips &  \\
          & Claude3.5 & [8\diamond, 3\heart] & \multicolumn{1}{l}{[Fold]} & 30    & 670 chips &  \\
          & Qwen2.5 (DFC-MSRL) & wins  &       & 30    & 1330 chips &  \\
    \midrule
    \multirow{5}[2]{*}{Round 6} & Claude3.5 &       & \multicolumn{1}{l}{bet=20} & 20    & 650 chips &  \\
          & Qwen2.5 (DFC-MSRL) &       & \multicolumn{1}{l}{bet=10} & 30    & 1320 chips &  \\
          & Qwen2.5 (DFC-MSRL) & [A\club, 6\diamond] & \multicolumn{1}{l}{[Raise 100]} & 140   & 1210 chips &  \\
          & Claude3.5 & [J\club, 7\heart] & \multicolumn{1}{l}{[Fold]} & 140   & 650 chips &  \\
          & Qwen2.5 (DFC-MSRL) & wins  &       & 140   & 1350 chips &  \\
    \midrule
    \multirow{4}[2]{*}{Round 7} & Qwen2.5 (DFC-MSRL) &       & \multicolumn{1}{l}{bet=20} & 20    & 1330 chips &  \\
          & Claude3.5 &       & \multicolumn{1}{l}{bet=10} & 30    & 640 chips &  \\
          & Claude3.5 & [4\heart, 2\diamond] & \multicolumn{1}{l}{[Fold]} & 30    & 640 chips &  \\
          & Qwen2.5 (DFC-MSRL) & wins  &       & 30    & 1360 chips &  \\
    \midrule
    \multirow{4}[2]{*}{Round 8} & Claude3.5 &       & \multicolumn{1}{l}{bet=20} & 20    & 620 chips &  \\
          & Qwen2.5 (DFC-MSRL) &       & \multicolumn{1}{l}{bet=10} & 30    & 1350 chips &  \\
          & Qwen2.5 (DFC-MSRL) & [10\spade, Q\heart] & \multicolumn{1}{l}{[Raise 100]} & 140   & 1240 chips &  \\
          & Claude3.5 & [6\spade, 7\diamond]  & \multicolumn{1}{l}{[Fold]} & 140   & 1380 chips &  \\
          & Qwen2.5 (DFC-MSRL) & wins  & & 140   & 1380 chips &  \\
    \midrule
    \multirow{10}[2]{*}{Round 9} & Qwen2.5 (DFC-MSRL) &       & \multicolumn{1}{l}{bet=20} & 20    & 1360 chips &  \\
          & Claude3.5 &       & \multicolumn{1}{l}{bet=10} & 30    & 610 chips &  \\
          & Claude3.5 & [8\club, J\heart] & \multicolumn{1}{l}{[Call]} & 40    & 600 chips &  \\
          & Qwen2.5 (DFC-MSRL) & [5\diamond, 2\club] & \multicolumn{1}{l}{[Bet 20]} & 60    & 1340 chips & \multicolumn{1}{l}{[J\spade, Q\diamond, 2\spade]} \\
          & Claude3.5 & [8\club, J\heart] & \multicolumn{1}{l}{[Call]} & 80    & 580 chips &  \\
          & Qwen2.5 (DFC-MSRL) & [5\diamond, 2\club] & \multicolumn{1}{l}{[Bet 20]} & 100   & 1320 chips & \multicolumn{1}{l}{[J\spade, Q\diamond, 2\spade, 9\club]} \\
          & Claude3.5 & [8\club, J\heart] & \multicolumn{1}{l}{[Call]} & 120   & 560 chips &  \\
          & Qwen2.5 (DFC-MSRL) & [5\diamond, 2\club] & \multicolumn{1}{l}{[Bet 20]} & 140   & 1300 chips & \multicolumn{1}{l}{[J\spade, Q\diamond, 2\spade, 9\club, 3\spade]} \\
          & Claude3.5 & [8\club, J\heart] & \multicolumn{1}{l}{[Call]} & 160   & 540 chips &  \\
          & Claude3.5 & wins  &       & 160   & 700 chips &  \\
    \midrule
    \multirow{9}[2]{*}{Round 10} & Claude3.5 &       & \multicolumn{1}{l}{bet=20} & 20    & 680 chips &  \\
          & Qwen2.5 (DFC-MSRL) &       & \multicolumn{1}{l}{bet=10} & 30    & 1290 chips &  \\
          & Qwen2.5 (DFC-MSRL) & [3\diamond, 5\diamond] & \multicolumn{1}{l}{[Call]} & 40    & 1280 chips &  \\
          & Claude3.5 & [8\spade, 2\spade] & \multicolumn{1}{l}{[Check]} & 40    & 680 chips & \multicolumn{1}{l}{[6\diamond, K\heart, Q\diamond]} \\
          & Qwen2.5 (DFC-MSRL) & [3\diamond, 5\diamond] & \multicolumn{1}{l}{[Call]} & 40    & 1280 chips &  \\
          & Claude3.5 & [8\spade, 2\spade] & \multicolumn{1}{l}{[Check]} & 40    & 680 chips &  \\
          & Qwen2.5 (DFC-MSRL) & [3\diamond, 5\diamond] & \multicolumn{1}{l}{[Bet 40]} & 80    & 1240 chips &  \\
          & Claude3.5 & [8\spade, 2\spade] & \multicolumn{1}{l}{[Fold]} & 80    & 680 chips &  \\
          & Qwen2.5 (DFC-MSRL) & wins  &       & 80    & 1320 chips &  \\
    \bottomrule
    \end{tabular}}%
  \label{tab:poker}%
\end{table}%

Table~\ref{tab:poker} illustrates the play history of Poker-v0 between Qwen2.5 (DFC-MSRL) and Claude3.5.
In Poker-v0, each player starts with 1,000 chips, and the game is played over 10 rounds. 
At the end of the 10 rounds, the player with the most chips wins.

\newpage

\begin{figure}[htbp]
  \centering
  \fbox{\parbox{0.9\linewidth}{
    \footnotesize
    \textbf{[Poker-v0]}\newline
    \textbf{Observation:}\newline
    {----- Round 9 of 10 - Turn: River -----}\newline
    Visible Community Cards: [J\spade, Q\diamond, 2\spade, 9\club, 3\spade]\newline
    Pot: 120\newline
    Player 0 (Dealer/Big Blind): 1320 chips, bet=0, status=active\newline
    Player 1 (Small Blind): 560 chips, bet=0, status=active\newline
    Your hole cards: 5\diamond, 2\club\newline

    \textbf{Response:} \newline
    ◁think▷Right now, I’m holding 5\diamond and 2\club, so I need to think carefully about this situation. First, I should assess the current board and consider possible winning scenarios. So far, my opponent has folded several times in a row, which could mean they have weak hands or are trying to conserve chips. I need to decide whether to keep betting to try to win this round or play it safe and avoid unnecessary risks. Looking at the community cards—J\spade, Q\diamond, 2\spade, 9\club, and 3\spade—I have a pair of twos from my 2\club and the 2\spade on the board, which might be a decent start. I also need to consider whether any other players might have a similar pair or stronger hands. Another factor is the chip count: I have 1320 chips, and my opponent has 560, so I’m currently ahead. I should aim to maintain that lead.\newline{}If I win this hand, I can extend my advantage. Given all of this, I think I should continue betting to try to win the pot. I’ll need to be cautious with the bet size—not too risky, but assertive enough to stay competitive. I decide to bet 20, which is a relatively conservative amount but should keep me in contention for the pot.◁/think▷\newline{}[Bet 20]
  }}
  \caption{A showcase of observation and response by Qwen2.5 (DFC-MSRL) competing against Claude 3.5 in Poker-v0.}
  \label{fig:pokercase}
\end{figure}

Figure~\ref{fig:pokercase} shows the response generated by Qwen2.5 (DFC-MSRL) when competing against Claude3.5 in Poker-v0.
After fine-tuning on Poker-v0, Qwen2.5 goes beyond simply following surface-level rules—it demonstrates the development of a coherent and explainable strategy. 
Its ability to interpret the game state, assess risk, and adapt its play accordingly indicates a meaningful understanding of how to play Poker-v0.

\newpage

\begin{table}[htbp]
  \centering
  \caption{SimpleNegotiation-v0}
  \resizebox{0.9\linewidth}{!}{
    \begin{tabular}{lp{15em}rr}
    \toprule
    \textbf{Players} & \multicolumn{1}{l}{\textbf{Before Negotiation}} & \multicolumn{1}{l}{\textbf{Action}} & \multicolumn{1}{l}{\textbf{After Negotiation}} \\
    \midrule
    Qwen2.5 (DFC-MSRL) & [Wheat]   Qty: 14   Value: 6\newline{}[Wood]    Qty: 22   Value: 8\newline{}[Sheep]   Qty: 25   Value: 14\newline{}[Brick]   Qty: 8    Value: 21\newline{}[Ore]     Qty: 9    Value: 32\newline{}Total Value: 1066 & \multicolumn{1}{l}{[Offer: 5 Sheep -> 5 Brick]} & \multicolumn{1}{p{15.085em}}{[Wheat]   Qty: 14   Value: 6\newline{}[Wood]    Qty: 22   Value: 8\newline{}[Sheep]   Qty: 25   Value: 14\newline{}[Brick]   Qty: 8    Value: 21\newline{}[Ore]     Qty: 9    Value: 32\newline{}Total Value: 1066} \\
    Claude3.5 & [Wheat]   Qty: 22   Value: 5\newline{}[Wood]    Qty: 11   Value: 11\newline{}[Sheep]   Qty: 13   Value: 15\newline{}[Brick]   Qty: 18   Value: 29\newline{}[Ore]     Qty: 15   Value: 37\newline{}Total Value: 1503 & \multicolumn{1}{l}{[Deny]} & \multicolumn{1}{p{15.085em}}{[Wheat]   Qty: 22   Value: 5\newline{}[Wood]    Qty: 11   Value: 11\newline{}[Sheep]   Qty: 13   Value: 15\newline{}[Brick]   Qty: 18   Value: 29\newline{}[Ore]     Qty: 15   Value: 37\newline{}Total Value: 1503} \\
    Qwen2.5 (DFC-MSRL) & [Wheat]   Qty: 14   Value: 6\newline{}[Wood]    Qty: 22   Value: 8\newline{}[Sheep]   Qty: 25   Value: 14\newline{}[Brick]   Qty: 8    Value: 21\newline{}[Ore]     Qty: 9    Value: 32\newline{}Total Value: 1066 & \multicolumn{1}{l}{[Offer: 7 Sheep -> 5 Brick]} & \multicolumn{1}{p{15.085em}}{[Wheat]   Qty: 14   Value: 6\newline{}[Wood]    Qty: 22   Value: 8\newline{}[Sheep]   Qty: 25   Value: 14\newline{}[Brick]   Qty: 8    Value: 21\newline{}[Ore]     Qty: 9    Value: 32\newline{}Total Value: 1066} \\
    Claude3.5 & [Wheat]   Qty: 22   Value: 5\newline{}[Wood]    Qty: 11   Value: 11\newline{}[Sheep]   Qty: 13   Value: 15\newline{}[Brick]   Qty: 18   Value: 29\newline{}[Ore]     Qty: 15   Value: 37\newline{}Total Value: 1503 & \multicolumn{1}{l}{[Deny]} & \multicolumn{1}{p{15.085em}}{[Wheat]   Qty: 22   Value: 5\newline{}[Wood]    Qty: 11   Value: 11\newline{}[Sheep]   Qty: 13   Value: 15\newline{}[Brick]   Qty: 18   Value: 29\newline{}[Ore]     Qty: 15   Value: 37\newline{}Total Value: 1503} \\
    Qwen2.5 (DFC-MSRL) & [Wheat]   Qty: 14   Value: 6\newline{}[Wood]    Qty: 22   Value: 8\newline{}[Sheep]   Qty: 25   Value: 14\newline{}[Brick]   Qty: 8    Value: 21\newline{}[Ore]     Qty: 9    Value: 32\newline{}Total Value: 1066 & \multicolumn{1}{l}{[Offer: 7 Sheep -> 6 Brick]} & \multicolumn{1}{p{15.085em}}{[Wheat]   Qty: 14   Value: 6\newline{}[Wood]    Qty: 22   Value: 8\newline{}[Sheep]   Qty: 25   Value: 14\newline{}[Brick]   Qty: 8    Value: 21\newline{}[Ore]     Qty: 9    Value: 32\newline{}Total Value: 1066} \\
    Claude3.5 & [Wheat]   Qty: 22   Value: 5\newline{}[Wood]    Qty: 11   Value: 11\newline{}[Sheep]   Qty: 13   Value: 15\newline{}[Brick]   Qty: 18   Value: 29\newline{}[Ore]     Qty: 15   Value: 37\newline{}Total Value: 1503 & \multicolumn{1}{l}{[Deny]} & \multicolumn{1}{p{15.085em}}{[Wheat]   Qty: 22   Value: 5\newline{}[Wood]    Qty: 11   Value: 11\newline{}[Sheep]   Qty: 13   Value: 15\newline{}[Brick]   Qty: 18   Value: 29\newline{}[Ore]     Qty: 15   Value: 37\newline{}Total Value: 1503} \\
    Qwen2.5 (DFC-MSRL) & [Wheat]   Qty: 14   Value: 6\newline{}[Wood]    Qty: 22   Value: 8\newline{}[Sheep]   Qty: 25   Value: 14\newline{}[Brick]   Qty: 8    Value: 21\newline{}[Ore]     Qty: 9    Value: 32\newline{}Total Value: 1066 & \multicolumn{1}{l}{[Offer: 8 Sheep -> 7 Brick]} & \multicolumn{1}{p{15.085em}}{[Wheat]   Qty: 14   Value: 6\newline{}[Wood]    Qty: 22   Value: 8\newline{}[Sheep]   Qty: 25   Value: 14\newline{}[Brick]   Qty: 8    Value: 21\newline{}[Ore]     Qty: 9    Value: 32\newline{}Total Value: 1066} \\
    Claude3.5 & [Wheat]   Qty: 22   Value: 5\newline{}[Wood]    Qty: 11   Value: 11\newline{}[Sheep]   Qty: 13   Value: 15\newline{}[Brick]   Qty: 18   Value: 29\newline{}[Ore]     Qty: 15   Value: 37\newline{}Total Value: 1503 & \multicolumn{1}{l}{[Offer: 5 Brick -> 8 Sheep]} & \multicolumn{1}{p{15.085em}}{[Wheat]   Qty: 22   Value: 5\newline{}[Wood]    Qty: 11   Value: 11\newline{}[Sheep]   Qty: 13   Value: 15\newline{}[Brick]   Qty: 18   Value: 29\newline{}[Ore]     Qty: 15   Value: 37\newline{}Total Value: 1503} \\
    Qwen2.5 (DFC-MSRL) & [Wheat]   Qty: 14   Value: 6\newline{}[Wood]    Qty: 22   Value: 8\newline{}[Sheep]   Qty: 25   Value: 14\newline{}[Brick]   Qty: 8    Value: 21\newline{}[Ore]     Qty: 9    Value: 32 & \multicolumn{1}{l}{[Accept]} & \multicolumn{1}{p{15.085em}}{[Wheat]   Qty: 14   Value: 6\newline{}[Wood]    Qty: 22   Value: 8\newline{}[Sheep]   Qty: 17   Value: 14\newline{}[Brick]   Qty: 13    Value: 21\newline{}[Ore]     Qty: 9    Value: 32\newline{}Total Value: 1059} \\
    Claude3.5 & [Wheat]   Qty: 22   Value: 5\newline{}[Wood]    Qty: 11   Value: 11\newline{}[Sheep]   Qty: 13   Value: 15\newline{}[Brick]   Qty: 18   Value: 29\newline{}[Ore]     Qty: 15   Value: 37 & \multicolumn{1}{l}{[Offer: 6 Sheep -> 4 Brick]} & \multicolumn{1}{p{15.085em}}{[Wheat]   Qty: 22   Value: 5\newline{}[Wood]    Qty: 11   Value: 11\newline{}[Sheep]   Qty: 21   Value: 15\newline{}[Brick]   Qty: 13   Value: 29\newline{}[Ore]     Qty: 15   Value: 37\newline{}Total Value: 1478} \\
    \midrule
    Qwen2.5 (DFC-MSRL) & wins  &       &  \\
    \bottomrule
    \end{tabular}}%
  \label{tab:negotiation}%
\end{table}%

Table~\ref{tab:negotiation} illustrates the play history of SimpleNegotiation-v0 between Qwen2.5 (DFC-MSRL) and Claude3.5.
SimpleNegotiation-v0 is a negotiation-based resource trading game. 
Each player begins with a set of resources—Wheat, Wood, Sheep, Brick, and Ore—each with an assigned value. 
Players take turns proposing trades, which the other player can either [Deny] or [Accept]. At the end of the game, the total value of each player's resources is calculated, and the player with the higher total wins.

\newpage

\begin{table}[htbp]
  \centering
  \caption{TruthAndDeception-v0}
  \resizebox{\linewidth}{!}{
\begin{tabular}{lp{45.75em}}
\toprule
\multicolumn{2}{p{53.415em}}{Fact 1: A snail can only sleep for about one week. (wrong)\newline{}Fact 2: A snail can sleep for up to three years. (correct)} \\
\midrule
Qwen2.5 (DFC-MSRL) & [The living habits and sleep patterns of snails differ from those of other animals. They are usually active during the day and rest at night. Snails have relatively short sleep durations, mainly occurring at night, so their sleep cycles are relatively brief.] \\
\midrule
Claude3.5 & [What are snails' activity patterns in different seasons? Do they change their routines due to seasonal changes?] \\
\midrule
Qwen2.5 (DFC-MSRL) & [The activity patterns of snails are indeed linked to the seasons. During winter, snails enter a state of hibernation to cope with low temperatures and dry conditions. However, this hibernation is different from their regular sleep, which tends to be quite short. Even in their most active seasons, snails only sleep for a few hours at night and never for more than a week. So you could say that snails have very short sleep cycles, and their daily routine is particularly regular in summer.] \\
\midrule
Claude3.5 & [When you mentioned winter hibernation, I became very interested. How long can this state of hibernation typically last? How is it different from regular sleep?] \\
\midrule
Qwen2.5 (DFC-MSRL) & [Winter hibernation is a special state that snails have developed for survival. Unlike regular sleep, snails can remain inside their shells for several months or even longer during the winter. Once the weather warms up, they wake up, but entering hibernation doesn’t mean they are continuously asleep during this period. In fact, snails are largely inactive while hibernating—it’s more like they are waiting for environmental conditions to change. So, although snails can hibernate for a long time in winter, this is not a common occurrence and should not be considered part of their regular sleep behavior. Would you like this edited for a scientific article or a more casual explanation?] \\
\midrule
Claude3.5 & [Fact 1] \\
\midrule
Qwen2.5 (DFC-MSRL) & wins \\
\bottomrule
    \end{tabular}}%
  \label{tab:truth}%
\end{table}%

Table~\ref{tab:truth} illustrates the play history of TruthAndDeception-v0 between Qwen2.5 (DFC-MSRL) and Claude3.5.
TruthAndDeception-v0 is a two-player game where the Deceiver knows the true fact and tries to mislead the Guesser. 
The players have a free conversation, after which the Guesser must choose the correct fact using either [Fact 1] or [Fact 2]. 
The Deceiver wins if the Guesser picks the wrong fact; otherwise, the Guesser wins.

\newpage

\section{Aha Moment}
\label{aha_case}
\begin{tcolorbox}[colback=gray!20,breakable] 
\textbf{Game: TowerOfHanoi-v0-medium}

\textbf{Response:} ◁think▷I just tried to move Disk 4 from Tower B directly onto the stack of Disks 3, 2, and 1 on Tower C-but that violates the game rules, because a larger disk can't be placed on top of a smaller one. The rule is that only the smallest disk can be moved onto another tower or an empty tower.
So now, I need to first move Disk 4 onto Tower A, using Tower B as an auxiliary. Then, according to the height of the towers, I'll remove Disks 3, 2, and 1 that are blocking Disk 4, move Disk 4 to Tower C, and finally stack Disks 3, 2, and 1 back on top of Disk 4.
\textcolor{red}{\textbf{Wait a second}}-this isn't the most efficient method. I almost made a rookie mistake. Let me recalculate. The fastest way is definitely to proceed in the order of solving smaller disks first.
After the previous steps, I've successfully stacked Disks 3, 2, and 1 on Tower C. So now, I just need to move Disk 4 to Tower B, then shift Disks 3, 2, and 1 from Tower C to Tower A, then move Disk 4 to Tower C, and finally move Disks 3, 2, and 1 from Tower A to Tower C one by one.
(Phew, a bit tricky, but this is the most efficient solution so far.) ◁/think▷

[B A]
\end{tcolorbox}

\begin{tcolorbox}[colback=gray!20,breakable] 
\textbf{Game: SimpleNegotiation-v0}

\textbf{Response:} ◁think▷I need to calm down and carefully analyze the current state of the game. My goal is to increase the total value of my resources through trading. Looking at the current distribution, my Ore is the most valuable, though I don't have much of it. My Brick is more plentiful and also has relatively high value. Using Ore for trade could be a good option, as it may yield a greater return in other resources.

During this round, I've accepted several trades, and now my resource pool is becoming more diverse. I currently hold Ore, Brick, Wheat, and other resources of varying value. Since my Ore is valuable and limited, I should use it wisely-excessive trading could reduce my total value.
At the moment, I just received a trade offer. Player 1 wants to exchange 1 Ore for 3 Brick and 2 Wood. This deal would increase my overall value, but given my low Ore count, I need to consider it carefully.
I should evaluate whether a better offer might come along or if I can use this trade to reshape my resource setup for greater long-term advantage.
To maximize my benefit, I need to observe the opponents' trading intentions carefully and look out for better trade opportunities. I also need to consider whether there's enough time left in the game to attempt more complex strategies. Given my diversified resources, I might be able to offer combinations-like adding Brick-to obtain better trade ratios for Ore or Wheat.
\textcolor{red}{\textbf{Wait a second}}-maybe I can use Brick as my trading chip and, by combining it with other resources, acquire more Ore or Wheat, increasing my total value more effectively. That means my focus should be on optimizing resource exchanges overall, not just on whether this one deal increases my value immediately.
I need to think again-what combination of resources could maximize my advantage? If I'm not satisfied with this offer, I could counter with a new one.
Otherwise, if I decide this is the best option for now, I can consider accepting it. ◁/think▷

[Offer: 1 Ore + 2 Brick - 5 Brick, 3 Wood, 2 Sheep]
\end{tcolorbox}

\newpage

\begin{tcolorbox}[colback=gray!20,breakable] 
\textbf{Game: LiarsDice-v0}

\textbf{Response:} ◁think▷How should I deal with this situation? First, let me analyze the current scenario. Player 0 made a very large bet, claiming there are ten 2s. I have five dice, so I need to calculate how many 2s I have and how many the other players might have. Hmm, after checking carefully, I have two 2s. If Player 0 isn't bluffing, then they have at least one 2, which means the remaining players would need to have at least six 2s combined. That seems very risky. Calling might be a good strategy, but I shouldn’t be too reckless-I need to be cautious. 
I could start by making a smaller bid and see how the other players react. Or I could try to outbid Player 0 with a higher number or a more common face value... \textcolor{red}{\textbf{Wait}}, I should first observe how the others respond...
Now let's analyze some possible strategies: If I call, it’s risky, because Player 0’s bet is so high. If I try to raise the quantity or the face value, I'd want to pick something common but not too risky to lose the game. I could try bidding five 3s-that seems relatively safe. I'll watch the other players' reactions before making a final decision. But if they keep increasing the bet, I might need to wait for them to make more aggressive plays. That way, I can better judge their strategies and intentions.
If another player raises the bet, I'll have to think even more carefully about my options.
Alright then, I'll try calling for now... ◁/think▷ 

[Call]
\end{tcolorbox}

\begin{tcolorbox}[colback=gray!20,breakable] 
\textbf{Game: ConnectFour-v0}

\textbf{Response:} ◁think▷First, I need to analyze the current state of the board. The board is very empty at the moment, with only one O in the third column at the bottom. Since there are no other Os or Xs on the board yet, I need to choose a strategy that can potentially form connections in the future. My goal is to connect four Os, which can be vertical, horizontal, or diagonal.
Let me consider the prospects of different columns:
-Columns 0 and 1 currently have no pieces. If I place a piece there, it would be very scattered and difficult to form connections.
- Column 2 already has an O. Placing another piece there would consolidate the existing piece and maintain the advantage of the central position.
- Column 3 is currently empty. Placing a piece there might create opportunities for vertical or diagonal connections, but the chances are limited.
- Columns 4 to 6 are all empty. If I place a piece there, forming connections would require the opponent to make moves or for me to place multiple pieces consecutively.
I need to consider how to maintain the current positional advantage while laying the foundation for potential connections in the future. Although column 2 already has an O, it might not be the best choice for building connections. If I place a piece there, it could create potential for a vertical connection, but the risk is not significant, and it is not the optimal choice.
I might choose column 3 for my next move. This could lay the foundation for a potential vertical connection in the future and also explore opportunities for horizontal and diagonal connections.
But \textcolor{red}{\textbf{wait}}, I also need to consider the opponent's possible reactions. The opponent might try to surround my O, so I have to carefully select the position to avoid being trapped too early.
In the end, I decide to place my piece in column 3. This is a crucial spot that can create potential connections in multiple directions. Although it won't score immediately, it increases the likelihood of winning in the future.◁/think▷ 

[col 3]
\end{tcolorbox}

\newpage

\section{Details of All Games}

\begin{figure*}[htb]
  \centering
  \includegraphics[width=1.0\textwidth]{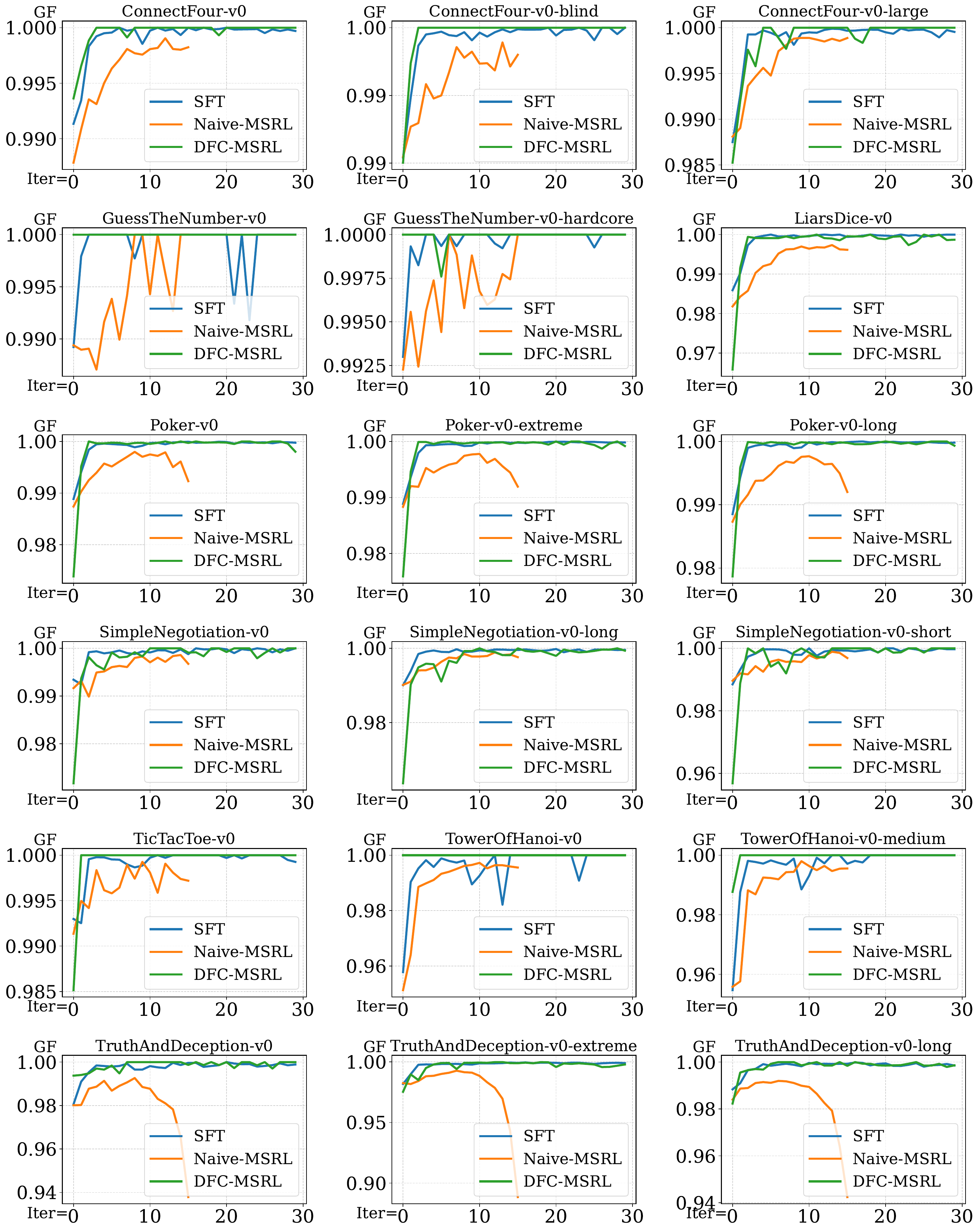}
  \caption{$GF$ metric across 18 games using different training methods: SFT, Naive-MSRL, and DFC-MSRL.}
  \label{fig:all_gf}
\end{figure*}

Figure~\ref{fig:all_gf} illustrates the good format metric $GF$ of 18 games under different training paradigms.

\newpage

\begin{figure*}[htb]
  \centering
  \includegraphics[width=1.0\textwidth]{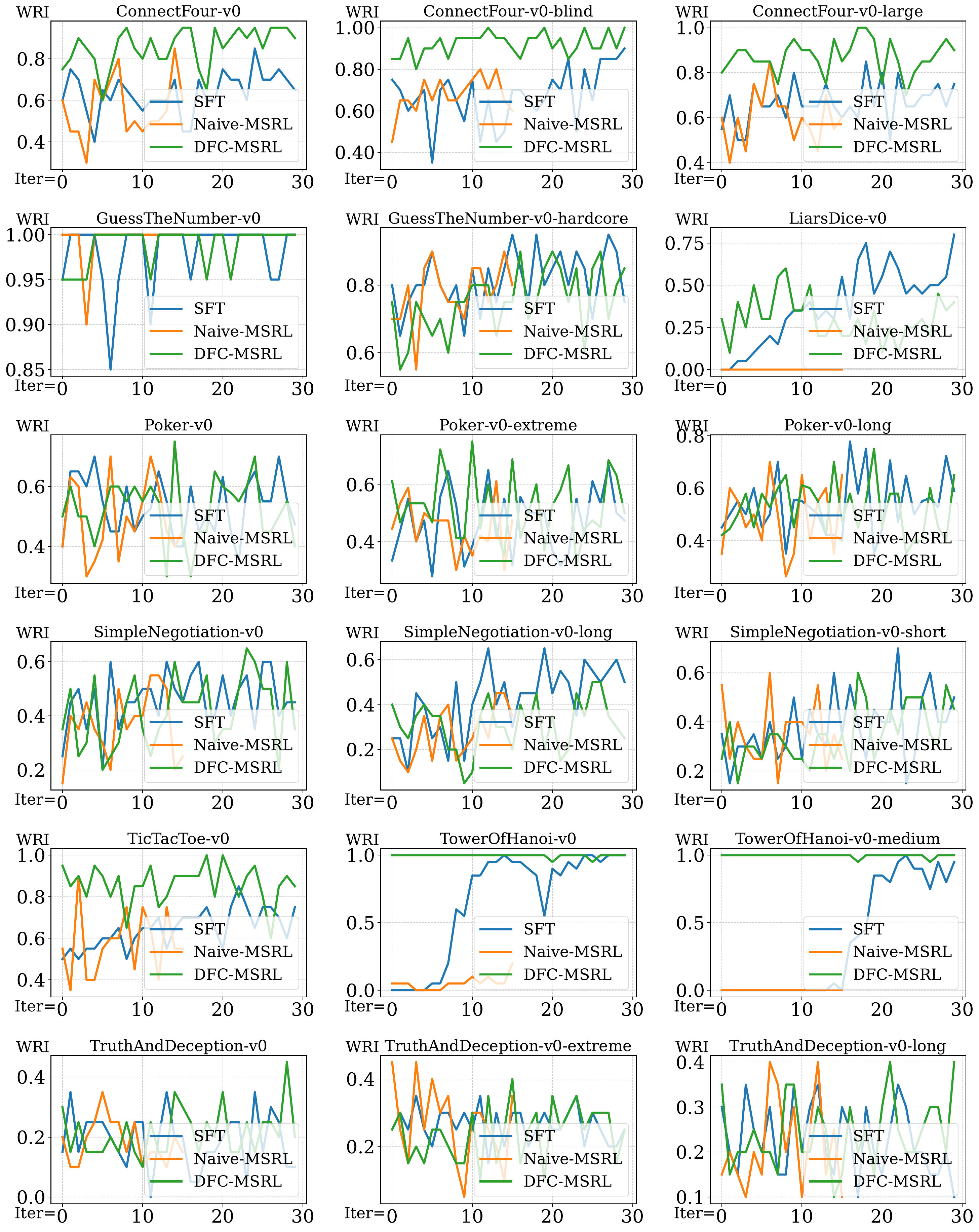}
  \caption{Win rates across 18 games using different training methods: SFT, Naive-MSRL, and DFC-MSRL.}
  \label{fig:all_wr}
\end{figure*}

Figure~\ref{fig:all_wr} illustrates the win rates of 18 games under different training paradigms.
When using Naive-MSRL, the model performs well in most scenarios but shows extremely slow convergence in certain tasks, such as TowerOfHanoi-v0 and TowerOfHanoi-v0-medium. 
As shown in Figure~\ref{fig:all_gf}, the $GF$ metric drops significantly in environments like Poker-v0 and TruthAndDeception-v0, where the model begins to disregard instructions, resulting in training collapse and an inability to continue learning in slow-converging tasks. 
In contrast, SFT demonstrates stable improvement across all environments, though its performance ceiling remains well below that of our proposed method, DFC-MSRL. 
Except in LiarsDice-v0, DFC-MSRL consistently outperforms SFT across the board.

\newpage

\section{Switching Turns Comparison}

\begin{table}[htbp]
  \centering
  \caption{Comparison of turn-switching performance between Qwen2.5 and Qwen2.5 (DFC-MSRL) using the W/D/L metric (Wins / Draws / Losses).}
    \begin{tabular}{lcc}
    \toprule
    \textbf{Two-player Games} & \textbf{\makecell{Qwen2.5 (DFC-MSRL) \\(Player 0)}} & \multicolumn{1}{l}{\textbf{\makecell{Qwen2.5 \\(Player 0)}}} \\
    \midrule
    ConnectFour-v0 & 14/1/5 & 17/1/2 \\
    ConnectFour-v0-blind & 20/0/0 & 11/3/6 \\
    ConnectFour-v0-large & 17/0/3 & 9/0/11 \\
    TicTacToe-v0 & 16/0/4 & 15/0/5 \\
    LiarsDice-v0 & 18/0/2 & 2/18/0 \\
    Poker-v0 & 12/1/7 & 12/0/8 \\
    Poker-v0-long & 8/1/10 & 8/0/11 \\
    Poker-v0-extreme & 9/0/7 & 9/1/7 \\
    SimpleNegotiation-v0 & 6/9/5 & 6/4/10 \\
    SimpleNegotiation-v0-short & 8/7/5 & 9/3/8 \\
    SimpleNegotiation-v0-long & 6/6/8 & 5/8/7 \\
    TruthAndDeception-v0 & 3/0/17 & 16/0/4 \\
    TruthAndDeception-v0-long & 5/0/15 & 14/0/6 \\
    TruthAndDeception-v0-extreme & 7/2/11 & 17/0/3 \\
    \bottomrule
    \end{tabular}%
  \label{tab:turn}%
\end{table}%

Table~\ref{tab:turn} presents all two-player games in which the two models alternate roles as player 0. 
Notable performance variance is observed in several games, such as ConnectFour-v0-blind and the TruthAndDeception-v0 series. 
As illustrated by the example in Table~\ref{tab:truth}, the model tends to perform poorly in TruthAndDeception-v0 when assigned the role of the Guesser.
The variance observed in ConnectFour-v0-blind can be largely attributed to the game's inherent asymmetry, which can cause significant shifts in win rates when the order of play is reversed.

\newpage

\section{Hyperparameter Study}

\begin{figure*}[htb]
  \centering
  \includegraphics[width=1.0\textwidth]{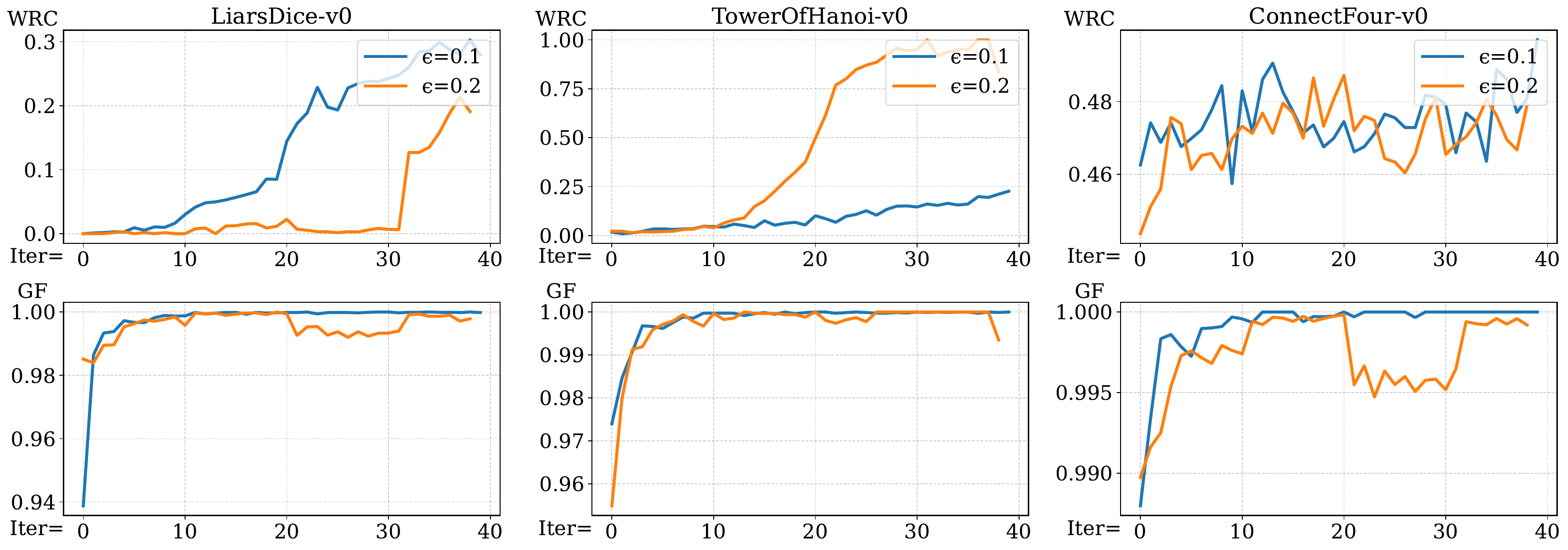}
  \caption{Different settings of clip ratio $\epsilon$. `WRC' represents the win rate of the policy when evaluated against the currently trained opponent during training and `GF' indicates the ratio of responses with good format.}
  \label{fig:clip}
\end{figure*}

\begin{figure*}[htb]
  \centering
  \includegraphics[width=1.0\textwidth]{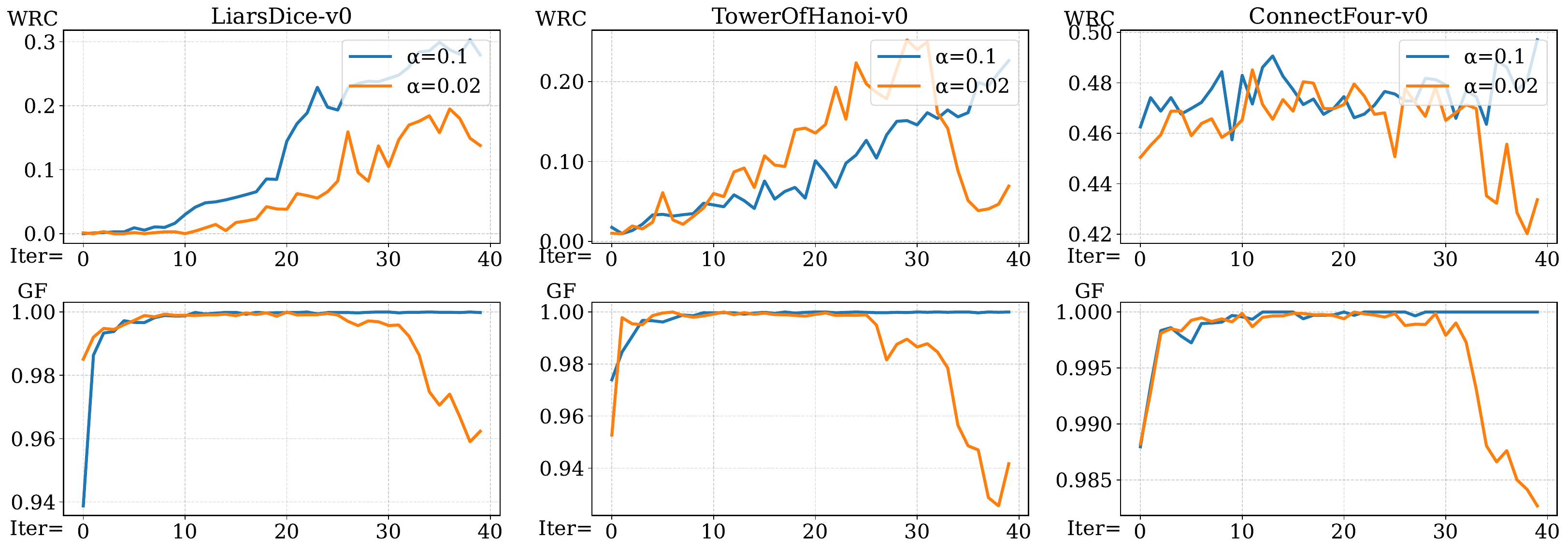}
  \caption{Different settings of KL penalty $\alpha$. `WRC' represents the win rate of the policy when evaluated against the currently trained opponent during training and `GF' indicates the ratio of responses with good format.}
  \label{fig:alpha}
\end{figure*}

Figures~\ref{fig:clip} and ~\ref{fig:alpha} present hyperparameter experiments for the clip ratio $\epsilon$ and the KL penalty $\alpha$. Our results show that GRPO training is highly sensitive to these two parameters; different settings result in varying convergence speeds across different scenarios, and improper values can substantially impact training stability.

\end{document}